%% file: main.tex
\pdfoutput=1
\documentclass{article}

\usepackage[utf8]{inputenc}
\usepackage[T1]{fontenc}
\usepackage{lmodern}
\usepackage[semibold,scale=0.96]{sourcesanspro}
\usepackage{inconsolata}
\usepackage{microtype}
\usepackage[a4paper, margin=2.5cm]{geometry}

\input{math_commands.tex}

\usepackage{amsmath, amssymb}
\usepackage{tikz}
\usetikzlibrary{arrows.meta,positioning,decorations.pathreplacing,calc}
\usepackage{float}
\usepackage{algorithm}
\usepackage{algpseudocode}
\usepackage[table,dvipsnames]{xcolor}
\PassOptionsToPackage{hypertexnames=false}{hyperref}
\usepackage[colorlinks]{hyperref}
\usepackage{url}
\usepackage{booktabs}
\usepackage{makecell}
\usepackage{wrapfig}
\usepackage{capt-of}
\usepackage{enumitem}
\usepackage{graphicx}
\usepackage{fancyhdr}
\usepackage{placeins}
\usepackage[round]{natbib}
\newcommand{\gv}[1]{#1}
\definecolor{bestpurple}{RGB}{106,61,154}
\newcommand{\bv}[1]{\textcolor{bestpurple}{\textbf{#1}}}


\definecolor{utorange}{HTML}{BF5700}
\definecolor{hqlaccent}{HTML}{5B2C8A}
\definecolor{hqldim}{HTML}{686868}
\definecolor{hqldot}{HTML}{CCC6C0}
\hypersetup{citecolor=hqlaccent, linkcolor=hqlaccent, urlcolor=hqlaccent}

\makeatletter
\newcommand{\toptitlebar}{{\color{black}\hrule height 1pt}\vskip 0.25in}
\renewcommand{\maketitle}{%
  \begin{center}%
    \toptitlebar
    \vskip 0.1in%
    {\LARGE\bfseries \@title \par}%
    \vskip 0.3in%
    {\large \@author \par}%
  \end{center}%
  \par
  \vskip 0.3in%
}
\renewcommand\section{\@startsection{section}{1}{\z@}{-2.0ex plus -0.5ex minus -.2ex}%
  {1.5ex plus 0.3ex minus .2ex}{\large\bfseries\raggedright}}
\renewcommand\subsection{\@startsection{subsection}{2}{\z@}{-1.8ex plus -0.5ex minus -.2ex}%
  {0.8ex plus .2ex}{\normalsize\bfseries\raggedright}}
\renewcommand\subsubsection{\@startsection{subsubsection}{3}{\z@}{-1.5ex plus -0.5ex minus -.2ex}%
  {0.5ex plus .2ex}{\normalsize\bfseries\raggedright}}
\renewcommand\paragraph{\@startsection{paragraph}{4}{\z@}{1ex plus .2ex minus .1ex}{-.5em}%
  {\normalfont\normalsize\bfseries}}
\renewenvironment{abstract}%
  {\centerline{\large\bfseries Abstract}%
   \begin{list}{}{\setlength{\rightmargin}{0.6cm}\setlength{\leftmargin}{0.6cm}}%
    \item[]\ignorespaces}%
  {\unskip\end{list}}
\makeatother

\newcommand{\utwordmark}{{\fontfamily{ptm}\selectfont\color{utorange}\fontsize{15}{15}\selectfont
  The University of Texas at Austin}}

\title{Horizon-Scalable Q-Learning}
\newcommand{\hqllinks}{{\sffamily\bfseries\fontsize{11.5}{14}\selectfont
  \href{https://ryanxhr.github.io/hql/}{\textcolor{hqlaccent}{Website}}%
  \hspace{1.1em}{\color{hqldot}\raisebox{0.18ex}{\scalebox{0.8}{$\bullet$}}}\hspace{1.1em}%
  \href{https://github.com/ryanxhr/hql}{\textcolor{hqlaccent}{GitHub}}\par}}

\newcommand{\paperauthors}{%
  {\fontsize{10.95}{15}\selectfont
  \mbox{\textbf{Haoran Xu}}\quad
  \mbox{\textbf{Pragya Srivastava}}\quad
  \mbox{\textbf{Xinjie Liu}}\quad
  \mbox{\textbf{Siddhant Agarwal}}\quad
  \mbox{\textbf{Amy Zhang}}\par}
  \vspace{6pt}
  {\normalsize\color{hqldim} The University of Texas at Austin\par}
  \vspace{3pt}
  {\small\ttfamily\color{hqldim} haoran.xu@utexas.edu\par}
  \vspace{13pt}
  \hqllinks
}
\author{\paperauthors}
\begin{document}

\maketitle
\thispagestyle{fancy}
\fancyhead{}
\fancyfoot[C]{\thepage}
\lhead{\raisebox{-0.55cm}{\utwordmark}}
\renewcommand{\headrulewidth}{0pt}
\setlength{\headheight}{29pt}
\setlength{\headsep}{3mm}

\begin{abstract}
Learning accurate off-policy state-action value functions (Q-functions) from offline datasets becomes harder as the task horizon grows. One-step temporal-difference (TD) learning bootstraps after every transition, so small per-step errors accumulate along the horizon, and early states can be badly misvalued even when the TD loss is small. Multi-step TD shortens this chain by following the dataset trajectory for several steps; however, since the behavior policy is often suboptimal, every followed action adds a bias, and no fixed number of steps suits a whole trajectory. In this paper, we propose Horizon-Scalable Q-Learning (HQL), a simple method that enables accurate long-horizon value learning by automatically balancing between one-step and multi-step returns. HQL extends a target through a dataset action only when the critic values it at least as highly as the best-of-N actions sampled from the BC policy, and otherwise falls back to a one-step target. We show that this rule is a principled off-policy correction that needs no action densities, which makes HQL compatible with expressive flow and diffusion policies. Experiments show that HQL avoids the failures of both one-step and multi-step TD on a combination-lock task across horizons, and outperforms prior horizon-reduction methods on long-horizon OGBench and OCBench tasks.
\end{abstract}

\section{Introduction} Off-policy reinforcement learning (RL) learns the state-action value function (Q-function) of a target policy from an offline dataset collected by a different behavior policy, which enables RL to reuse logged experience in real-world domains such as robotics \citep{levine2016end, kalashnikov2018qtopt} and industrial control \citep{zhan2022deepthermal}. However, off-policy value learning becomes harder as the \emph{effective horizon} grows \citep{laidlaw2023effective}, i.e., the number of transitions across which the critic must carry value information before reaching a state whose value it already knows. 
Most off-policy actor-critic methods learn the critic with one-step temporal-difference (TD) backups \citep{fujimoto2019bcq, kumar2020cql, kostrikov2022iql, xu2023ivr, park2025fql}. A one-step target bootstraps after a single logged transition, which makes it robust to off-policy data but moves value back by only one transition per update. We show that the value error at a state is exactly the sum of the one-step Bellman residuals along the rest of the trajectory, so a critic can reach a small TD loss and still misvalue the start states by an amount that grows with the horizon.

Multi-step returns \citep{watkins1989learning, sutton2018book} or Q-chunking \citep{li2025qc, zhao2023act} shorten this chain by following the dataset trajectory for several steps before bootstrapping. Nevertheless, this imposes a trade-off between fast value propagation (longer backups) and low off-policy bias (shorter backups): we show that each spanned suboptimal action adds a systematic bias that grows toward a limit linear in $H$. Since offline trajectories could be mixed with suboptimal actions, no fixed backup length is optimal for every TD update. Per-decision off-policy corrections such as Retrace \citep{precup2000eligibility, munos2016retrace} make this choice at every transition with importance ratios. However, these ratios require policy densities, which flow or diffusion-based policies \citep{park2025fql, mao2024diffusiondice, li2026qam, xu2026vgf} do not expose.

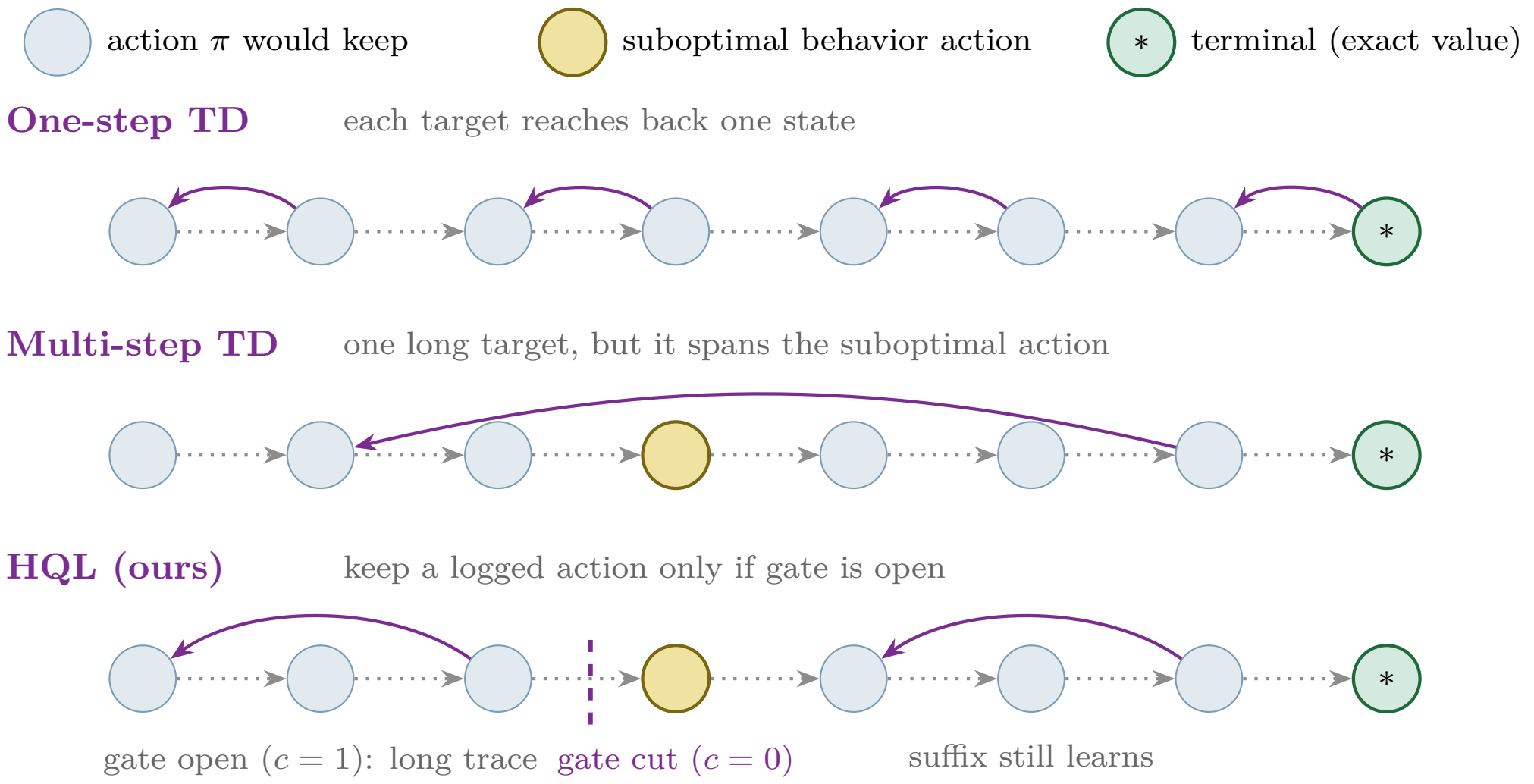
\begin{figure}[t]
\centering
\definecolor{hqlpurple}{RGB}{123,44,145}
\definecolor{hqlblue}{RGB}{49,104,142}
\definecolor{hqlgreen}{RGB}{44,150,88}
\definecolor{hqlyellow}{RGB}{220,185,30}
\begin{tikzpicture}[
    >={Stealth[length=2mm]},
    state/.style={circle,draw,minimum size=5.8mm,inner sep=0pt,font=\footnotesize},
    good/.style={state,fill=hqlblue!14,draw=hqlblue!65},
    bad/.style={state,fill=hqlyellow!42,draw=hqlyellow!55!black,thick},
    term/.style={state,fill=hqlgreen!20,draw=hqlgreen!70!black,thick},
    chain/.style={->,dotted,thick,black!45},
    span/.style={->,thick,hqlpurple},
    gate/.style={hqlpurple,dashed,line width=1.1pt},
    rowhead/.style={font=\small\bfseries,text=hqlpurple,anchor=west},
    rowsub/.style={font=\footnotesize,text=black!60,anchor=west},
    note/.style={font=\footnotesize,text=black!60,anchor=north},
    lbl/.style={font=\footnotesize,anchor=west},
    xscale=1.55,
]

\path (-0.85,2.05) -- (7.85,2.05);
\node[good] at (-0.48,2.05) {};
\node[lbl] at (-0.28,2.05) {action $\pi$ would keep};
\node[bad] at (2.42,2.05) {};
\node[lbl] at (2.62,2.05) {suboptimal behavior action};
\node[term] at (5.62,2.05) {$\ast$};
\node[lbl] at (5.82,2.05) {terminal (exact value)};

\node[rowhead] at (-0.85,1.35) {One-step TD};
\node[rowsub] at (1.05,1.35) {each target reaches back one state};
\foreach \i in {0,...,6} {\node[good] (a\i) at (\i,0.4) {};}
\node[term] (a7) at (7,0.4) {$\ast$};
\foreach \i/\j in {0/1,1/2,2/3,3/4,4/5,5/6,6/7} {\draw[chain] (a\i)--(a\j);}
\foreach \i/\j in {1/0,3/2,5/4,7/6} {\draw[span] (a\i) to[bend right=55] (a\j);}

\begin{scope}[yshift=-1.95cm]
\node[rowhead] at (-0.85,1.35) {Multi-step TD};
\node[rowsub] at (1.05,1.35) {one long target, but it spans the suboptimal action};
\foreach \i in {0,1,2} {\node[good] (b\i) at (\i,0.4) {};}
\node[bad] (b3) at (3,0.4) {};
\foreach \i in {4,5,6} {\node[good] (b\i) at (\i,0.4) {};}
\node[term] (b7) at (7,0.4) {$\ast$};
\foreach \i/\j in {0/1,1/2,2/3,3/4,4/5,5/6,6/7} {\draw[chain] (b\i)--(b\j);}
\draw[span] (b6) to[bend right=20] (b1);
\end{scope}

\begin{scope}[yshift=-3.9cm]
\node[rowhead] at (-0.85,1.35) {HQL (ours)};
\node[rowsub] at (1.05,1.35) {keep a logged action only if gate is open};
\foreach \i in {0,1,2} {\node[good] (c\i) at (\i,0.4) {};}
\node[bad] (c3) at (3,0.4) {};
\foreach \i in {4,5,6} {\node[good] (c\i) at (\i,0.4) {};}
\node[term] (c7) at (7,0.4) {$\ast$};
\foreach \i/\j in {0/1,1/2,2/3,3/4,4/5,5/6,6/7} {\draw[chain] (c\i)--(c\j);}
\draw[span] (c2) to[bend right=48] (c0);
\draw[span] (c6) to[bend right=48] (c4);
\draw[gate] (2.52,0.0) -- (2.52,0.82);
\node[note] at (1.0,-0.05) {gate open ($c=1$): long trace};
\node[note,text=hqlpurple] at (3.0,-0.05) {gate cut ($c=0$)};
\node[note] at (5.0,-0.05) {suffix still learns};
\end{scope}

\end{tikzpicture}
\caption{\textbf{Horizon-Scalable Q-Learning (HQL).} Purple arcs show the reach of one training target. One-step TD accumulates residuals, and multi-step TD inherits the bias of the suboptimal action it spans (yellow). HQL compares each logged action with the best-of-$N$ policy samples, continuing the trace where it passes ($c{=}1$) and cutting it where it fails ($c{=}0$, dashed), so reward travels far without inheriting off-policy bias.}
\label{fig:intuition}
\end{figure}

In this paper, we propose Horizon-Scalable Q-Learning (HQL), which shortens the effective horizon of Q-learning without inheriting the bias of suboptimal dataset actions. HQL achieves this by automatically balancing between one-step and multi-step returns at every transition (Figure~\ref{fig:intuition}). More specifically, HQL attaches a binary \emph{gate} to every logged action, i.e., every action stored in the dataset: if the target critic values it at least as highly as the best-of-$N$ actions sampled from the BC policy, the gate is open and the target continues through it for one more step; otherwise, the target falls back to a one-step bootstrap. Our key observation is that this gate is principled without any density: for the optimal policy $\pi\propto\mu\exp(\tau Q)$ that behavior-regularized RL approximates \citep{peters2010reps, peng2019awr, xu2025unirl}, the behavior density cancels from the importance ratio, and as $\tau\to\infty$ the largest gate that preserves $Q^\pi$ becomes exactly the comparison of critic values. HQL uses an estimated behavior model only to sample action candidates and never evaluates its density. Since the gate depends only on the ordering of critic values, it is also invariant to the reward scale, and the only hyperparameter in HQL is the number of candidates $N$. 

We evaluate HQL on a combination-lock task, whose true $Q$-values are known exactly, and on long-horizon OGBench and OCBench tasks \citep{park2025ogbench,park2026ocbench}. On the combination lock, HQL keeps the lowest value estimation error across different horizon, whereas one-step TD accumulates error and $n$-step TD fails on corrupted dataset. On OGBench and OCBench, at a matched budget of critic targets, HQL outperforms one-step TD, $n$-step TD, Q-chunking \citep{li2025qc}, and Long-horizon Q-Learning \citep{abraham2026lql}, especially on the most challenging long-horizon tasks. Perhaps surprisingly, we find that most of HQL's targets are one or few-step targets: HQL wins by choosing \emph{which} targets to extend, not by extending them far.

\section{Related Work}
\paragraph{Error propagation in value learning.} Error propagation through bootstrapping has been studied for approximate value and policy iteration \citep{munos2005error, farahmand2010error}, TD with function approximation \citep{tsitsiklis1997analysis}, and $n$-step and TD($\lambda$) updates \citep{kearns2000bias}; its instability with off-policy data is the deadly triad \citep{vanhasselt2018deadly}, and the horizon governs the difficulty of evaluation and learning \citep{liu2018breaking, laidlaw2023effective}. 

\paragraph{Multi-step off-policy correction.} Multi-step returns and eligibility traces shorten the backup chain \citep{watkins1989learning, sutton2018book, deasis2018multistep}, and a small fixed $n$ is standard in deep RL \citep{hessel2018rainbow, kapturowski2019r2d2, hernandez2019understanding}. Per-decision importance sampling \citep{precup2000eligibility} and its clipped variants (Retrace \citep{munos2016retrace, touati2018convergent}, off-policy $Q(\lambda)$ \citep{harutyunyan2016qlambda}, V-trace \citep{espeholt2018impala}, trajectory-aware traces \citep{daley2023trajectory}) correct the mismatch step by step, and emphatic TD reweights updates \citep{sutton2016emphatic}. Watkins's $Q(\lambda)$ cuts traces at non-greedy actions while Peng's $Q(\lambda)$ keeps them \citep{peng1996incremental, kozuno2021peng, kim2026peng}; other work adapts the trace \citep{white2016greedy, rowland2020adaptive}, avoids ratios via action-dependent bootstrapping \citep{mahmood2017abq}, gates traces at non-greedy actions \citep{daley2026gated}, or adds multi-step inequalities \citep{abraham2026lql}. However, these corrections need action probabilities or a discrete greedy target policy.

\paragraph{Horizon reduction.} Instead of correcting within a backup, horizon-reduction methods change what a decision is. Hierarchical methods act over options or subgoals \citep{bacon2017option, vezhnevets2017feudal, nachum2018hiro, nachum2019why, xu2022por}, and action chunking treats a short action sequence as one decision \citep{seo2025chunk, li2025qc, li2025decoupled, song2026chunk}. However, within each segment they use a fixed, uncorrected backup and inherit the bias; HQL is orthogonal and contains both as ungated special cases. 
Another line exploits the structure of goal-reaching tasks through quasimetric value functions \citep{wang2023quasimetric, zheng2025multistep}, graph search over subgoals \citep{eysenbach2019search}, and goal-conditioned hierarchies \citep{park2023hiql, ahn2025option, zhou2025flattening, park2025dual}. These methods rely on a goal-reaching reward and thus apply only to goal-conditioned RL; in contrast, HQL makes no assumption on the reward.

\paragraph{Policy extraction with expressive policies.} Offline RL methods regularize the policy toward the data, either explicitly through divergence penalties \citep{wu2019brac, kumar2019bear, xu2021sbac, fujimoto2021minimalist} or implicitly through in-sample value learning and weighted behavior cloning \citep{peng2019awr, nair2020awac, kostrikov2022iql, xu2022por, xu2023ivr, xu2025idrl}. Recent work extracts expressive diffusion or flow policies by value-guided training \citep{wang2023diffusionql, mao2024diffusiondice, park2025fql}, rejection sampling from a behavior model \citep{chen2023sfbc, hansenestruch2023idql}, adjoint matching \citep{li2026qam}, or value gradient flow \citep{xu2026vgf}. These methods keep a one-step critic; HQL only changes the critic's target and can serve any of them.

\section{Preliminaries}
\label{sec:pre}
We consider the problem presented as a finite-horizon Markov decision process (MDP) $\mathcal M=(\mathcal S,\mathcal A,P,r,H)$, where $\mathcal S$ and $\mathcal A$ are the state and action spaces, $P(s'|s,a)$ and $r(s,a)$ denote the transition dynamics and reward function, and $H$ is the horizon; an episode that reaches its goal early enters a zero-reward absorbing state, so every episode has length $H$. For a policy $\pi(a|s)$, the action value is $Q^\pi(s,a)=\mathbb E_\pi[\sum_t r(s_t,a_t)\mid s_0=s,a_0=a]$, with $V^\pi(s)=\mathbb E_{a\sim\pi(\cdot|s)}[Q^\pi(s,a)]$ and advantage $A^\pi(s,a)=Q^\pi(s,a)-V^\pi(s)$. We consider the off-policy setting, where one needs to estimate $Q^\pi$ of a target policy $\pi$ from a dataset (or replay buffer) $\mathcal D$ collected by a possibly suboptimal behavior policy $\mu(a|s)$. 

\subsection{Error Accumulation in One-Step and $n$-Step TD}
\label{sec:amplification}

\paragraph{Error decomposition along a chain.} To understand how critic error scales with the horizon, we derive one error law for one-step and $n$-step backups (derivations in Appendices~\ref{app:onestep}--\ref{app:nstep}). Consider the deterministic chain $s_0\rightarrow s_1\rightarrow\cdots\rightarrow s_H$ induced by a fixed deterministic target policy, with an exactly known terminal value. Write $Q_t=Q(s_t,a_t)$, let $r_t$ be the reward on the transition out of $s_t$, and let $\overline Q$ be the target critic, with $\overline V^\pi(s)=\mathbb{E}_{a\sim\pi(\cdot|s)}[\overline Q(s,a)]$. The $n$-step target is
\begin{equation}
    Y_t^{(n)}
    =
    \sum_{j=0}^{n-1}r_{t+j}
    +
    \overline V^\pi(s_{t+n}),
    \label{eq:nstep-target}
\end{equation}
where a true terminal state replaces the last term by its known value. The loss at lookahead $n$ is $L^{(n)}_{\mathrm{TD}}=\mathbb{E}_t[(Q_t-\operatorname{sg}(Y_t^{(n)}))^2]$, where $\operatorname{sg}$ stops gradients~\citep{sutton2018book}; for $n=1$ we write $L_{\mathrm{TD}}$. With the one-step Bellman residual $e_t=Q_t-r_t-Q_{t+1}$ and the value error $\epsilon_t=Q_t-Q_t^\pi$, we have $e_t=\epsilon_t-\epsilon_{t+1}$ because $Q_t^\pi=r_t+Q_{t+1}^\pi$, and with an exact terminal value ($\epsilon_H=0$) the recursion telescopes:
\begin{equation}
    \epsilon_t=\sum_{\tau=t}^{H-1}e_\tau.
    \label{eq:telescope}
\end{equation}
In other words, the value error is \emph{exactly} the sum of the downstream residuals, for any critic at any point in training.

\paragraph{A unified error law.} To see how the residuals add up, we model them as random noise across training runs: each residual has zero mean and variance $\sigma^2$, and residuals at different steps are only weakly correlated (Appendix~\ref{app:onestep}). Under this model, applying Eq.~\ref{eq:telescope} to $n$-step backups, one block of $n$ steps at a time, gives a single expression for the start-state error (Appendix~\ref{app:nstep}):
\begin{equation}
    \operatorname{RMSE}(\epsilon_0)^2
    \;\approx\;
    \underbrace{\frac{H}{n}\,L^{(n)}_{\mathrm{TD}}}_{\text{residual accumulation}}
    \;+\;
    \underbrace{\left(1-\tfrac1n\right)^{\!2} H^2\,\bar A_\mu^2}_{\text{behavior-action bias}},
    \label{eq:nstep-rmse}
\end{equation}
where $\bar A_\mu\le0$ is the average advantage of the logged actions that the targets span. Intuitively, the first term decreases with $n$ while the second increases with $n$, so a longer backup trades residual accumulation for behavior-action bias.

\paragraph{One-step TD: residuals accumulate as $\Theta(\sqrt H)$.} A one-step target spans no logged action, so the bias term vanishes, and summing Eq.~\ref{eq:telescope} under the residual model gives, as $H-t\to\infty$,
\begin{equation}
    \operatorname{RMSE}(\epsilon_t)=\sqrt{\mathbb{E}[\epsilon_t^2]}\sim
    \sigma_{\mathrm{eff}}\sqrt{H-t},
    \label{eq:rmse}
\end{equation}
with $\sigma_{\mathrm{eff}}^2=\sigma^2+2\sum_{\ell\ge1}\operatorname{Cov}(e_t,e_{t+\ell})$ the long-run variance of the residuals. This result is important as it shows that a small training loss does not certify a small value error: since $L_{\mathrm{TD}}=\mathbb{E}[e_t^2]=\sigma^2$ measures only the per-step residual, Eq.~\ref{eq:nstep-rmse} at $n=1$ reads
\begin{equation}
    \operatorname{RMSE}(\epsilon_0)=\sqrt{H\,L_{\mathrm{TD}}}
    \label{eq:one-step-loss-horizon}
\end{equation}
up to the factor $\sigma_{\mathrm{eff}}/\sigma$. Appendix~\ref{app:onestep} discusses when this rate fails and how a discount $\gamma<1$ caps it at a horizon of order $1/(1-\gamma)$.

\paragraph{$n$-step TD: bias grows toward $H|\bar A_\mu|$.}
\label{sec:nstep}
Bootstrapping $n$ steps ahead divides the residual term by $n$. However, along a logged trajectory $Y_t^{(n)}$ now deviates from $Q^\pi(s_t,a_t)$ by $\sum_{j=1}^{n-1}A^\pi(s_{t+j},a_{t+j})$ plus the bootstrap error at the endpoint (Eq.~\ref{eq:nstep-error-decomposition}). For a near-optimal target policy every spanned advantage is at most zero, so this bias is systematic and does not average out; as $n$ grows, the square root of the second term of Eq.~\ref{eq:nstep-rmse} approaches $H|\bar A_\mu|$. This indicates that no fixed $n$ suits all logged actions: the best lookahead $n$ changes from step to step, which calls for a per-transition rule to minimize the total error.

\section{Horizon-Scalable Q-Learning}
\label{sec:hql}

In this section, we introduce Horizon-Scalable Q-Learning (HQL). We begin with a gated multi-step target and design the gate based on value comparison (Section~\ref{sec:gate}). We then show that this gate is a limiting case of per-decision importance sampling yet needs no action density (Section~\ref{sec:retrace-connection}). %
Finally, we provide several practical considerations for HQL and discuss its connection to previous work (Section~\ref{sec:hwindow}). %

\subsection{Gated Multi-Step Targets}
\label{sec:gate}

Our key idea is to let a backup continue through a logged action only when that action is at least as good as the actions the target policy would take, i.e., its advantage under the target policy is non-negative, which keeps it out of the bias term of Eq.~\ref{eq:nstep-rmse} while reducing the horizon of TD learning via multi-step bootstrapping. 
To do so, we attach a gate $c_i\in\{0,1\}$  to every transition and define the HQL target backward along a logged segment,
\begin{equation}
    Y_t^{\textsc{hql}}
    =
    \underbrace{r_t+\gamma\,\overline V^\pi(s_{t+1})}_{\text{one-step target}}
    \;+\;
    \gamma\,c_{t+1}
    \underbrace{\big(Y_{t+1}^{\textsc{hql}}-\overline Q(s_{t+1},a_{t+1})\big)}_{\text{downstream correction}},
    \label{eq:retrace-target}
\end{equation}
with the correction set to zero at the end of the segment, and $\gamma$ kept for generality. Intuitively, the first term is the usual one-step bootstrap from the target policy, and the second term carries back whatever the logged continuation has learned beyond the critic's current estimate at $(s_{t+1},a_{t+1})$. When the gate accepts $a_{t+1}$, this correction flows back and the target extends one more step; when it rejects $a_{t+1}$, the target is a plain one-step target, and the states after the cut still start their own targets (as shown in Figure~\ref{fig:intuition}, bottom). Unrolling the recursion, a run of accepted actions yields a multi-step target whose length is set by the data rather than fixed in advance. Note that $c\equiv0$ recovers one-step TD and $c\equiv1$ is the uncorrected multi-step return.

It remains to decide when the gate is open, i.e., whether the target policy would keep the logged action. Intuitively, a logged action is safe to follow if it is at least as good as what the policy itself would propose, which the critic can check by comparing the logged action with candidates drawn from the policy's own sampler:
\begin{equation}
    c_i
    =
    \mathbb{I}\!\left[
        \overline Q(s_i,a_i)
        \;\ge\;
        \max_{1\le m\le N}\overline Q\big(s_i,a_i^{(m)}\big)
    \right],
    \qquad
    a_i^{(m)}\sim\mu(\cdot|s_i),
    \label{eq:bon-gate}
\end{equation}
where $\mu$ is the behavior policy that can be estimated using any advanced generative models from $\mathcal D$. In other words, the logged action is kept if it is at least as good as the best of $N$ policy proposals. Note that the gate depends only on an ordering of critic values, so it is invariant to the scale of $Q$, needs no behavior density, and has $N$ as its only hyperparameter; it costs $N$ behavior samples and critic evaluations per transition, the same as the best-of-$N$ bootstrap already requires.

\subsection{A Density-Free Gate via Per-Decision Importance Sampling}
\label{sec:retrace-connection}
A remaining question is whether cutting multi-step backups with this gate is principled, since cutting changes what a multi-step backup converges to. Per-decision importance sampling answers this \citep{precup2000eligibility, munos2016retrace}: in the tabular setting, any gate $c_i\in[0,1]$ with $c_i\le\pi(a_i|s_i)/\mu(a_i|s_i)$ preserves $Q^\pi$ as the fixed point of Eq.~\ref{eq:retrace-target}, and we call such a gate \emph{valid}. However, evaluating this bound requires the density ratio $\pi/\mu$, which is unavailable and hard to estimate \citep{mao2024odice, xu2025idrl} when using generative policies, or when policies are implicit, e.g.,  $\mu$ is a flow sampler and $\pi$ selects among its samples. Fortunately, the ratio has a closed form for the policy that behavior-regularized RL methods approximate, i.e., the exponential tilt of the behavior policy \citep{peters2010reps, peng2019awr, xu2023ivr, li2026qam},
\begin{equation}
    \pi_\theta(a|s)
    =
    \frac{
        \mu(a|s)\exp(\tau Q(s,a))
    }{
        Z(s)
    },
    \qquad
    Z(s)
    =
    \mathbb{E}_{a\sim\mu(\cdot|s)}
    \left[\exp(\tau Q(s,a))\right],
    \label{eq:qam-tilt}
\end{equation}
so the behavior density cancels and $\log(\pi/\mu)(a_i|s_i)=\tau Q(s_i,a_i)-\log Z(s_i)$. We estimate the normalizer from $N$ behavior samples as $\widehat{\log Z}(s_i)=\operatorname{LSE}_m(\tau\overline Q(s_i,a_i^{(m)}))-\log N$, where $\operatorname{LSE}_m(x_m)=\log\sum_{m=1}^{N}\exp(x_m)$ is the log-sum-exp, and clipping the ratio at one gives the largest valid gate
\begin{equation}
    c_i
    =
    \exp
    \left(
        \min\left\{
            0,\,
            \tau\overline Q(s_i,a_i)-\widehat{\log Z}(s_i)
        \right\}
    \right).
    \label{eq:qam-retrace-coefficient}
\end{equation}
This soft gate interpolates between two extremes. As $\tau\to0$, every gate equals one and the target is an uncorrected multi-step return. As $\tau\to\infty$, the tilt concentrates on the best behavior-supported action, and because $\operatorname{LSE}_m(x_m)\le\max_m x_m+\log N$, the exponent is nonnegative at any $\tau$ when no candidate beats the logged action and tends to $-\infty$ otherwise, so the soft gate becomes exactly the value comparison of Eq.~\ref{eq:bon-gate}. Therefore, HQL's gate is the $\tau\to\infty$ limit of a valid gate, which is more scalable and robust since it avoids a temperature and can be computed without density. It also outperforms the soft gate at every $\tau$ we tried (Appendix~\ref{app:softgate}).

\subsection{Practical Considerations}
\label{sec:hwindow}
\begin{wrapfigure}{r}{0.5\linewidth}
\centering
\vspace{-1.3em}
\definecolor{hqlpurple}{RGB}{123,44,145}
\definecolor{hqlblue}{RGB}{49,104,142}
\definecolor{hqlgreen}{RGB}{44,150,88}
\resizebox{\linewidth}{!}{%
\begin{tikzpicture}[
    >={Stealth[length=2mm]},
    st/.style={circle,draw,minimum size=4.4mm,inner sep=0pt,font=\tiny,
               fill=hqlblue!14,draw=hqlblue!65},
    dec/.style={rounded corners=1.2pt,draw=hqlgreen!70!black,fill=hqlgreen!10,thick},
    chain/.style={->,dotted,thick,black!45},
    span/.style={->,thick,hqlpurple},
    gate/.style={hqlpurple,dashed,line width=1.0pt},
    rowhead/.style={font=\normalsize\bfseries,text=hqlpurple,anchor=west},
    note/.style={font=\small,text=black!60,anchor=north west},
    xscale=1.10,
]
\node[rowhead] at (-0.55,1.22) {$L{=}1$,\ $H{=}4$: HQL};
\foreach \i in {0,...,7} {\node[st] (b\i) at (\i,0.60) {};}
\foreach \i in {0,...,7} {\draw[dec] (\i-0.26,0.34) rectangle (\i+0.26,0.86);}
\foreach \i/\j in {0/1,1/2,2/3,3/4,4/5,5/6,6/7} {\draw[chain] (b\i)--(b\j);}
\draw[span] (b3) to[bend right=22] (b0);
\draw[gate] (3.42,0.24) -- (3.42,0.96);
\node[note] at (-0.55,0.16) {gate open for $3$ decisions, then cut};

\begin{scope}[yshift=-2.0cm]
\node[rowhead] at (-0.55,1.22) {$L{=}2$,\ $H{=}1$: Q-chunking};
\foreach \i in {0,...,7} {\node[st] (c\i) at (\i,0.60) {};}
\foreach \i in {0,2,4,6} {\draw[dec] (\i-0.28,0.34) rectangle (\i+1.28,0.86);}
\foreach \i/\j in {0/1,1/2,2/3,3/4,4/5,5/6,6/7} {\draw[chain] (c\i)--(c\j);}
\draw[span] (c2) to[bend right=32] (c0);
\node[note] at (-0.55,0.16) {one chunk, nothing to gate};
\end{scope}

\begin{scope}[yshift=-4.0cm]
\node[rowhead] at (-0.55,1.22) {$L{=}2$,\ $H{=}4$: Chunked-HQL at stride $2$};
\foreach \i in {0,...,7} {\node[st] (d\i) at (\i,0.60) {};}
\foreach \i in {0,2,4,6} {\draw[dec] (\i-0.28,0.34) rectangle (\i+1.28,0.86);}
\foreach \i/\j in {0/1,1/2,2/3,3/4,4/5,5/6,6/7} {\draw[chain] (d\i)--(d\j);}
\draw[span] (d5) to[bend right=16] (d0);
\draw[gate] (5.45,0.24) -- (5.45,0.96);
\node[note] at (-0.55,0.16) {$3$ decisions $=6$ steps kept, then cut};
\end{scope}
\end{tikzpicture}}
\caption{\textbf{HQL and Chunked-HQL.} $H$ sets how many \emph{decisions} one target chains, and $L$ how many \emph{environment steps} each decision covers.}
\label{fig:stride-vs-h}
\end{wrapfigure}
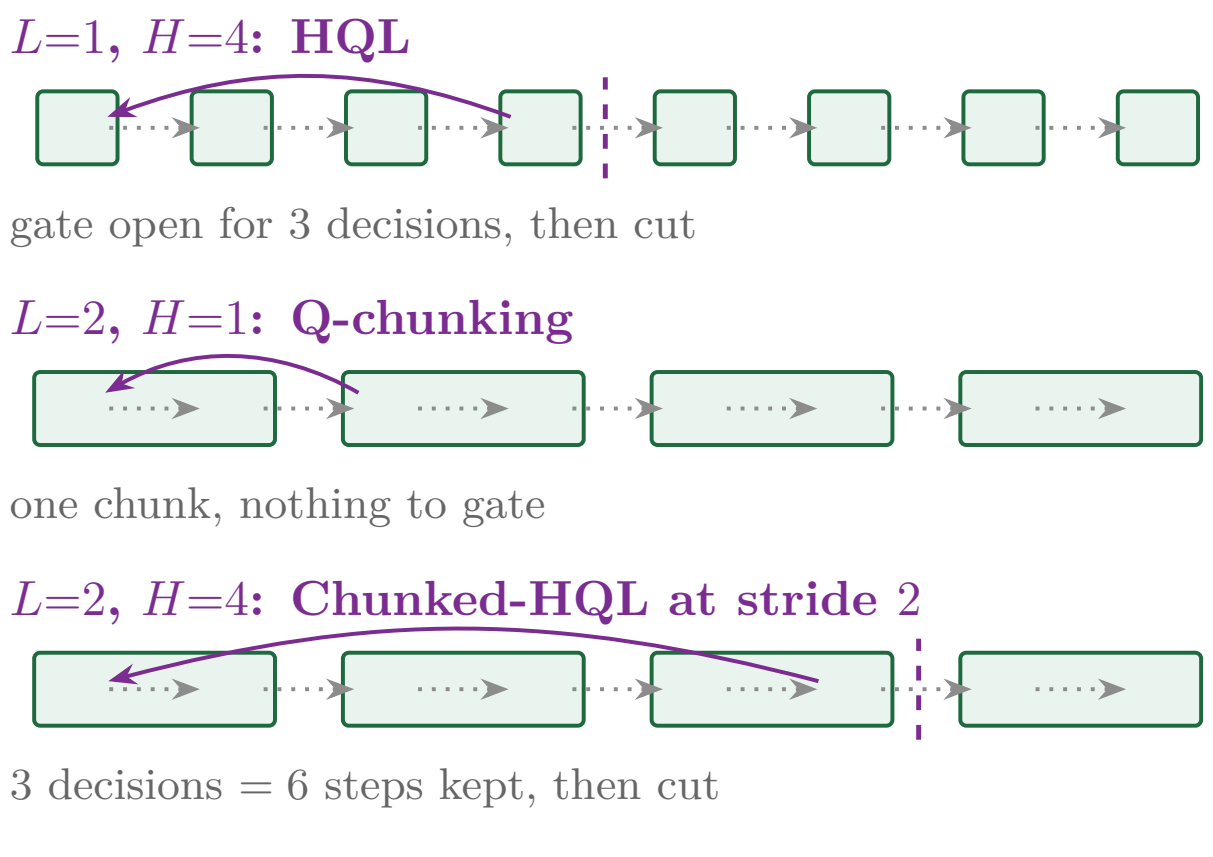
\paragraph{Dense targets over a trajectory window.} Since HQL is not restricted to a fixed $n$-step backup, it does not need to compute a target only at the first state of a sampled segment. Instead, HQL samples a trajectory window $(s_t,a_t,r_t,\dots,s_{t+H})$ of $H$ consecutive transitions and computes the gated multi-step target for \emph{every} state--action pair in the window. More specifically, it computes the gates $c_i$ once at every position, evaluates Eq.~\ref{eq:retrace-target} backward from the window end, and regresses each $(s_{t+j},a_{t+j})$ onto its own target $Y_{t+j}^{\textsc{hql}}$, $j=0,\dots,H-1$, masking positions beyond an episode boundary (Algorithm~\ref{alg:hql}). By doing so, a window yields $H$ targets whose maximal reach decreases from $H$ to $1$, and the gate can shorten each of them further. To enable a fair comparison, we hold the number of critic targets per update fixed: with a budget of $B$, one-step TD samples $B$ transitions and HQL samples $B/H$ windows. 

\begin{algorithm}[t]
\caption{Horizon-Scalable Q-Learning (critic update)}
\label{alg:hql}
\begin{algorithmic}[1]
\Require dataset $\mathcal D$, window length $H$, gate width $N$, target budget $B$
\State Sample $B/H$ windows $(s_t,a_t,r_t,\dots,s_{t+H})$ of $H$ consecutive transitions from $\mathcal D$
\State At every position $i$: draw $a_i^{(1:N)}\sim\mu(\cdot|s_i)$ and compute the gate $c_i$ (Eq.~\ref{eq:bon-gate})
\State For $j=H-1,\dots,0$: compute $Y_{t+j}^{\textsc{hql}}$ by Eq.~\ref{eq:retrace-target}, with no correction at $j=H-1$
\State Regress all $B$ pairs $(s_{t+j},a_{t+j})$ onto $\operatorname{sg}(Y_{t+j}^{\textsc{hql}})$; update $\overline Q$
\end{algorithmic}
\end{algorithm}

\paragraph{Multi-step decisions via Chunked-HQL.}
\label{sec:stride}
So far the decision stride is fixed to 1, meaning each decision corresponds to a single environment step. To allow for more flexible decision-making, we let one \emph{decision} span several environment steps. More specifically, with stride $L$, a decision is a chunk of $L$ primitive actions that the critic scores as a unit, $Q(s_t, a_{t:t+L})$, and a window of $H$ chunks reaches $L\!\cdot\!H$ environment steps while still producing $H$ targets. Note that the gate is unchanged, now comparing the logged chunk with the best-of-$N$ chunks drawn from the policy, and the backup of Eq.~\ref{eq:retrace-target} holds with $\gamma^L$ in place of $\gamma$ and per-chunk reward sums.

\paragraph{Connections to existing methods.} To summarize, in HQL or Chunked-HQL, $L$ sets how many environment steps one decision covers, and $H$ sets how many decisions one target may chain (Figure~\ref{fig:stride-vs-h}). With $L{=}1$, $H{=}1$ gives one-step TD, an ungated window of size $H$ gives uncorrected $H$-step TD, and a gated window gives HQL. With $L{=}k$, $H{=}1$ is Q-chunking with chunks of $k$ actions \citep{li2025qc}, since a single chunk leaves nothing to gate, and $H{>}1$ gives Chunked-HQL at stride $k$.
Note that the two variables extend the maximal reach of a target, $L\cdot H$ environment steps, in different ways. Increasing $H$ lets one target makes more decisions while the policy outputs less actions per step, and the gate decides for each target how many of these decisions it actually uses. In contrast, increasing $L$ makes each decision cover more environment steps, but each chunk then runs open loop and the policy must search a larger, $L\cdot\dim(\mathcal A)$-dimensional action space. 

\section{Experiments}

The goal of our experiments is to answer four questions. \textbf{(Q1)} Does HQL enable better value learning under different horizon lengths and dataset suboptimality? \textbf{(Q2)} How does HQL compare with TD and other baselines on OGBench and OCBench? \textbf{(Q3)} Does HQL work with different policy-extraction methods? \textbf{(Q4)} How do the different components of HQL affect its performance?%

\subsection{Experimental Setup}
\label{sec:setup}

\paragraph{Combination lock.} For Q1, we use a modified \texttt{combination-lock} chain \citep{park2025horizon} with H linearly ordered states and two actions: the ``correct'' one advances to the next state and the other returns to the first state, with reward -1 per step and 0 at the terminal state. The offline dataset consists of length-n=64 windows (64 transitions each): for every state--action pair we roll out M=100 windows that start with that pair and then follow the correct action, truncating a window early if it reaches the terminal state. Following \citet{park2025horizon}, we construct datasets of $n$-step trajectory windows with uniform coverage of starting state-action pairs. To study the effect of increasing action corruption, we independently replace each of the $n-1=63$ subsequent actions with the non-optimal action with probability $p \in \{0, 0.002, 0.005, 0.011, 0.022\}$, chosen so that in expectation $(1-p)^{63}\in\{100, 88, 73, 50, 25\}\%$ of windows are clean, i.e.\ contain no corrupted action. We use $H \in \{512,1024,2048,4096\}$, giving 2HM windows per dataset. Because the dynamics are deterministic, the ground-truth Q is available in closed form (\ref{app:combo-lock-q}).
\paragraph{OGBench.} Q2--Q4 use six state-based OGBench configurations \citep{park2025ogbench}: \texttt{cube-double}, \texttt{cube-triple}, \texttt{scene} (sparse reward), \texttt{humanoidmaze-medium}, \texttt{antmaze-giant} and \texttt{humanoidmaze-giant}, whose task horizons last $500$ to $4000$ steps (Table~\ref{tab:ogbench-tasks}); we report the mean score over each configuration's five tasks. Runs train for 1M offline and then 1M online steps, with $50$ evaluation episodes every $10^5$ steps; a run's score is the mean of its last three evaluations. 

\paragraph{OCBench.} To test HQL on even longer horizons, we also use OCBench \citep{park2026ocbench}, a robotic manipulation benchmark whose datasets come from scripted policies with properties similar to human demonstrations. We take eight full-frequency ($50$\,Hz) tasks from four domains (\texttt{block}, \texttt{chamber}, \texttt{switch}, \texttt{hanoi}), whose episodes last $1000$ to $5000$ steps. Following the RL baselines of OCBench, all methods use $25$-step action chunks, $8\times2048$ residual MLPs, an HL-Gauss critic and $\gamma{=}0.9999$, with $1024$ critic targets per update. Runs train for 1M offline and 1.5M online steps with $500$ evaluation episodes every $5\times10^4$ steps (Appendix~\ref{app:ocbench-setup}).

\paragraph{Baselines.} We compare HQL ($H=16$) with existing baselines under a matched compute budget: every method receives $512$ critic targets per gradient step. We consider three TD baselines. 16-window TD uses the same windows and batch as HQL but removes the gate, so comparing the two isolates the gate's effect. 8-step TD ($n{=}8$ matches the average reach of a $16$-step window) and 1-step TD each regress a single target per sequence with a batch of $512$.  
We also compare with Q-chunking (QC)~\citep{li2025qc}, which treats a chunk of $8$ actions as a single decision and bears no off-policy bias, and Long-horizon Q-Learning (LQL)~\citep{abraham2026lql}, which adds hinge penalties to one-step TD that enforce multi-step inequalities along the logged trajectory. All methods share the same critic architecture and use the same behavior flow policy with best-of-N sampling. 

\subsection{Q1: A Toy Example with Corrupted Windows}
\label{sec:toy}

\begin{figure}[t]
    \centering
    \includegraphics[width=1.0\linewidth]{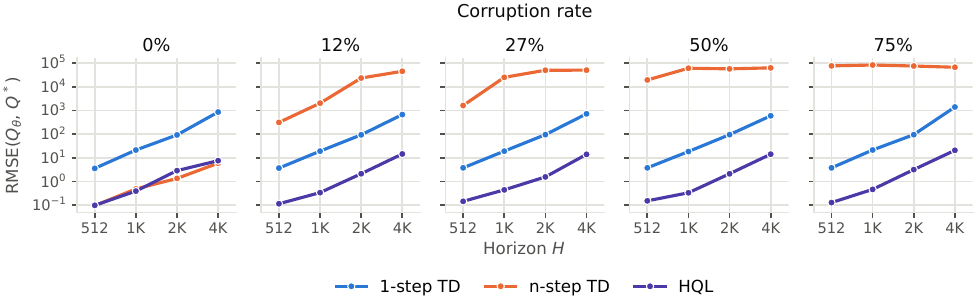}
    \caption{\textbf{Combination lock.} HQL remains accurate under data corruption as the horizon grows. The figure illustrates the RMSE between the learned $Q_\theta$ and the true $Q^*$, averaged uniformly over all $2H$ state–action pairs of the corrupted combination lock, after $1M$ updates (log scale). Panels vary the corruption fraction of the dataset. n-step TD ($n=64$) diverges as soon as any corruption is present; 1-step TD is unaffected by corruption but its error grows sharply with the horizon $H$.}
    \label{fig:combination_lock}
\end{figure}
\begin{figure}[t] %
    \centering
    \includegraphics[width=1.0\linewidth]{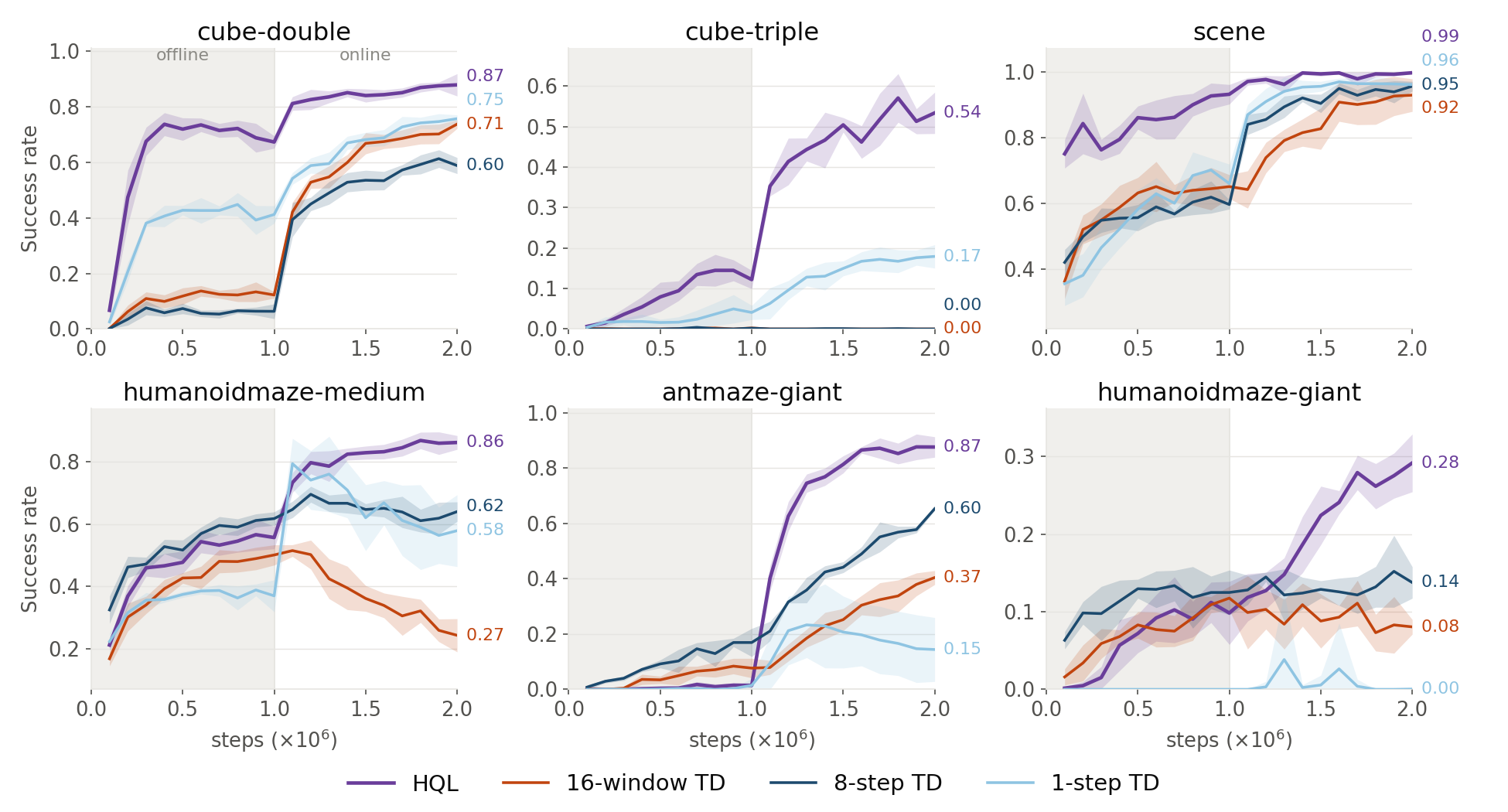}
    \caption{\textbf{HQL versus baseline TD methods at matched critic update budget.} Success rate averaged over the five tasks of each configuration and over $5$ seeds, with standard deviation shown as the band; gray marks the offline phase. HQL outperforms all baselines on all six configurations, especially on the challenging long-horizon ones.}
    \label{fig:h16-taskmean}
    \vspace{0.6em}
    \includegraphics[width=1.0\linewidth]{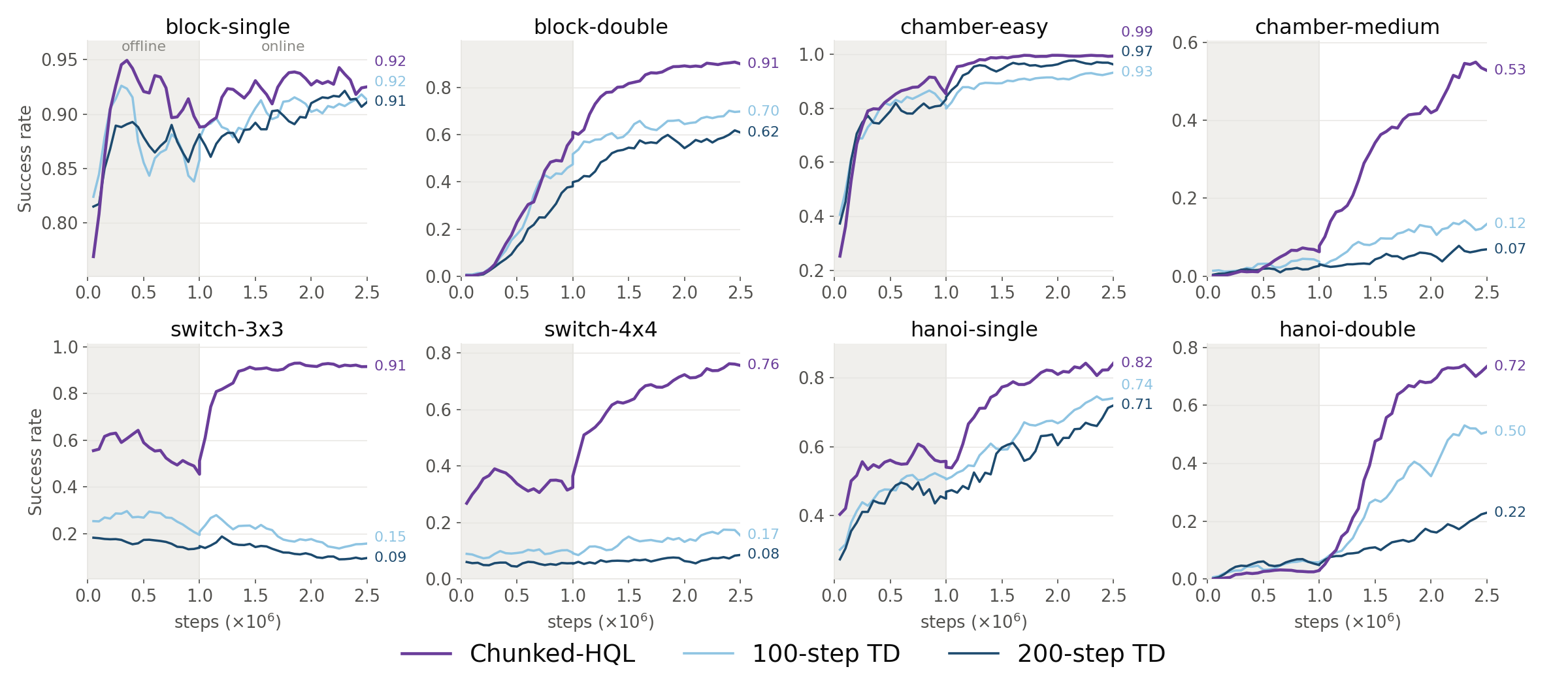}
    \caption{\textbf{Chunked-HQL versus $n$-step TD on OCBench.} Success rate on eight $50$\,Hz tasks (one seed, smoothed over three evaluations; numbers are the mean of the last three). All methods use $25$-step chunks and the same best-of-$16$ bootstrap. Chunked-HQL matches or outperforms both baselines on all eight tasks, especially on \texttt{switch} and \texttt{chamber-medium}.}
    \label{fig:ocbench}
\end{figure}

Figure~\ref{fig:combination_lock} supports both halves of the analysis. On clean data, the error of 1-step TD grows by three orders of magnitude from $H{=}512$ to $4096$, while 64-step TD and HQL avoid this accumulation. As the corruption rate increases, 64-step TD degrades sharply, because each target sums the negative advantages of the actions it spans. HQL cuts the trace at those actions: its error stays below both 1-step TD and 64-step TD at every horizon. 

\subsection{Q2: Results on OGBench and OCBench}
\label{sec:benchmark}

\paragraph{HQL is the strongest method across all six domains.} HQL reaches a mean success rate of 0.74 against 0.49 for the best TD baseline, 8-step TD (Figure~\ref{fig:h16-taskmean}). Consistent with our analysis in Section~\ref{sec:amplification}, the gap is largest on the four configurations with the largest horizons of 1000 steps or more (Table~\ref{tab:ogbench-tasks}); on \texttt{cube-triple}, HQL reaches 0.54 while both multi-step baselines score 0. 16-window TD matches HQL in everything except the gate, yet it is the weakest method on average, this demonstrates the effectiveness of gating in HQL.

\paragraph{Comparison with other baselines.} At the same budget and sampling width (Table~\ref{tab:more-baselines}), HQL improves the mean success rate by $12\%$ over LQL and by $29\%$ over QC. QC is best on \texttt{cube-double} and \texttt{scene} but scores $0\%$ on both humanoid mazes.

\begin{figure}[t]
\centering
\begin{minipage}[t]{0.345\linewidth}
\centering
\includegraphics[width=\linewidth]{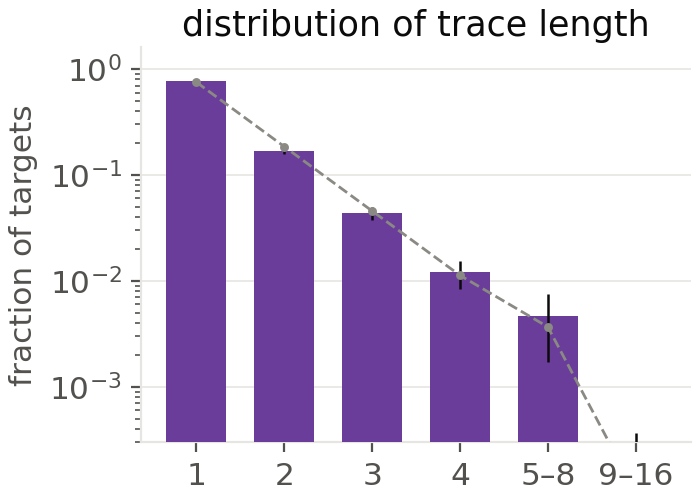}
\raggedright
\caption{\textbf{Per-target trace length} with the memoryless prediction (dashed).}
\label{fig:trace-hist}
\end{minipage}\hfill
\begin{minipage}[t]{0.63\linewidth}
\centering
\includegraphics[width=\linewidth]{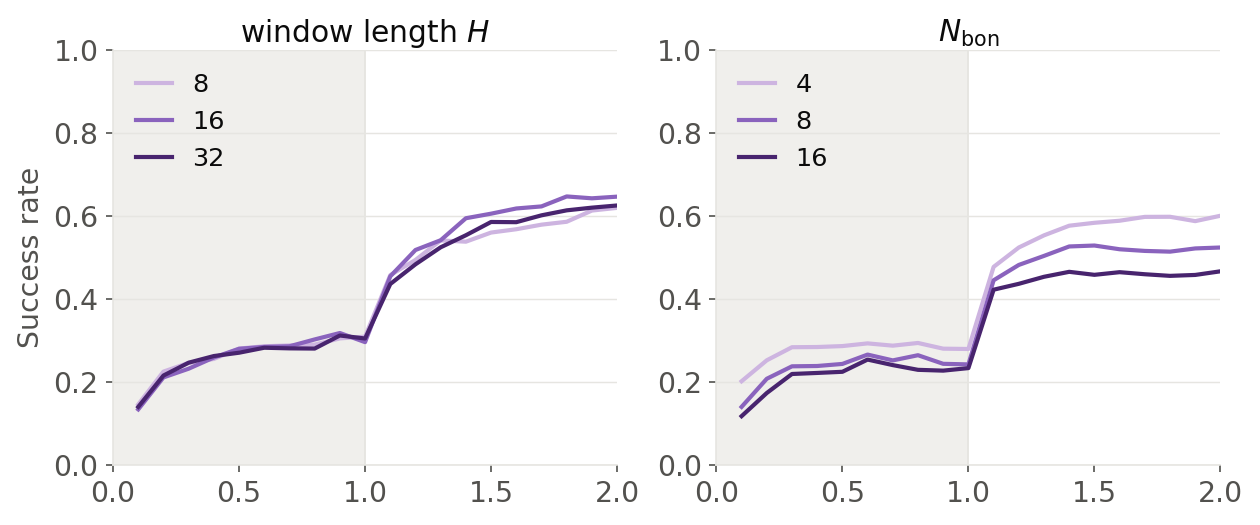}
\caption{\textbf{HQL ablations} over window length $H$ (left) and best-of-$N$ gate width $N$ (right). $H$ matters less than $N$, and a smaller $N$ gives better results.}
\label{fig:ablations}
\end{minipage}
\end{figure}

\begin{wraptable}{r}{0.57\linewidth}
\centering
\vspace{-1em}
\caption{\textbf{HQL and Chunked-HQL against LQL and QC.} Success rates (\%) are averaged over five tasks and reported as offline $\to$ online with standard error over five seeds; we bold the best offline and the best online score at integer level. HQL and Chunked-HQL achieve the best results, especially on the long-horizon tasks.}
\label{tab:more-baselines}
\vspace{0.2em}
\setlength{\tabcolsep}{2.2pt}
\scriptsize
\begin{tabular}{lcccc}
\toprule
& HQL & Chunked-HQL & LQL & QC \\
\midrule
\texttt{antmaze-giant} & \bv{1}\,$\to$\,\bv{87}{\tiny$\pm$2} & \gv{0}\,$\to$\,\gv{86}{\tiny$\pm$1} & \gv{0}\,$\to$\,\gv{44}{\tiny$\pm$4} & \gv{0}\,$\to$\,\gv{46}{\tiny$\pm$1} \\
\texttt{cube-double} & \gv{67}\,$\to$\,\gv{87}{\tiny$\pm$1} & \bv{73}\,$\to$\,\bv{100}{\tiny$\pm$0} & \gv{45}\,$\to$\,\gv{83}{\tiny$\pm$2} & \gv{28}\,$\to$\,\bv{100}{\tiny$\pm$0} \\
\texttt{cube-triple} & \bv{12}\,$\to$\,\gv{54}{\tiny$\pm$1} & \gv{7}\,$\to$\,\bv{84}{\tiny$\pm$1} & \gv{5}\,$\to$\,\gv{35}{\tiny$\pm$3} & \gv{1}\,$\to$\,\gv{26}{\tiny$\pm$3} \\
\texttt{hmaze-giant} & \gv{10}\,$\to$\,\gv{28}{\tiny$\pm$1} & \gv{0}\,$\to$\,\bv{29}{\tiny$\pm$2} & \bv{12}\,$\to$\,\gv{26}{\tiny$\pm$1} & \gv{0}\,$\to$\,\gv{0}{\tiny$\pm$0} \\
\texttt{hmaze-medium} & \gv{56}\,$\to$\,\gv{86}{\tiny$\pm$1} & \bv{68}\,$\to$\,\bv{88}{\tiny$\pm$2} & \gv{48}\,$\to$\,\gv{86}{\tiny$\pm$1} & \gv{0}\,$\to$\,\gv{0}{\tiny$\pm$0} \\
\texttt{scene} & \gv{93}\,$\to$\,\gv{99}{\tiny$\pm$0} & \bv{98}\,$\to$\,\bv{100}{\tiny$\pm$0} & \gv{96}\,$\to$\,\gv{97}{\tiny$\pm$1} & \gv{97}\,$\to$\,\bv{100}{\tiny$\pm$0} \\
\midrule
\textbf{mean} & \gv{40}\,$\to$\,\gv{74}{\tiny$\pm$1} & \bv{41}\,$\to$\,\bv{81}{\tiny$\pm$1} & \gv{34}\,$\to$\,\gv{62}{\tiny$\pm$1} & \gv{21}\,$\to$\,\gv{45}{\tiny$\pm$1} \\
\bottomrule
\end{tabular}
\vspace{-0.5em}
\end{wraptable}
Although Q-chunking avoids the off-policy bias by having the critic to score the whole L-sized action chunk $Q(s_t, a_{t:t+L})$, it falls short because the critic must choose among $L\cdot\mathrm{dim}(\mathcal A)$-dimensional chunks, and a fixed best-of-$N$ search covers that space increasingly sparsely as the chunk size $L$ grows. Second, each chunk runs open loop for $L$-steps (we choose $L=8$ while running QC), so the agent cannot correct its action within a chunk. This also explains why QC performs well on \texttt{cube-double} and \texttt{scene}, whose chunks are low-dimensional (please refer to table \ref{tab:ogbench-tasks} for action dimensions of the different OGBench tasks), but trails on \texttt{antmaze-giant} and fails on both humanoid mazes. We expect a wider best-of-$N$ search to help QC, since it covers the chunk space more densely, though at a cost in samples and critic evaluations that grows with $N$. The gate is also complementary to chunking: adding it on top of chunked actions (Chunked-HQL) raises QC's success rate from $45\%$ to $81\%$ (Table~\ref{tab:more-baselines}). 
\paragraph{Why does HQL help?} One might expect a multi-step method to win by sending reward information far back along the trajectory. To understand why HQL works better than the TD baselines, we log how many steps each target's trace survives (Figure~\ref{fig:trace-hist}). Perhaps surprisingly, the actual traces are short: $77\%$ of targets are exactly one-step targets while $17\%$ extend to two steps and $6\%$ to three or more steps, and none exceeds six of the sixteen available steps. This indicates that HQL helps not by making targets longer but by choosing \emph{which} ones to extend: $n$-step TD extends all of them and inherits the bias of every action spanned, 1-step TD extends none, and HQL extends a target only when its logged action passes the gate.

\paragraph{Chunked-HQL scales to the longer horizons of OCBench.} On OCBench, we compare Chunked-HQL ($L{=}25$, $H{=}8$, i.e., a $200$-step window) with $n$-step TD at $n\in\{100,200\}$, which regresses the chunk critic onto an uncorrected $n$-step return and is the backup used by OCBench's own RL baselines ($n{=}200$). Motivated by Section~\ref{sec:ablations}, we decouple the two widths of HQL: the gate uses $N{=}2$, while the bootstrap and acting use best-of-$16$ as in the baselines. Chunked-HQL reaches a mean success rate of $0.82$ at $2.5$M steps against $0.53$ for 100-step TD and $0.46$ for 200-step TD (Figure~\ref{fig:ocbench}). The gap is largest on \texttt{switch-3x3} ($0.91$ vs.\ $0.15$), \texttt{switch-4x4} ($0.76$ vs.\ $0.17$) and \texttt{chamber-medium} ($0.53$ vs.\ $0.12$), where the $n$-step baselines barely improve or even degrade during online training. Note that 200-step TD is worse than 100-step TD on seven of the eight tasks, which agrees with Eq.~\ref{eq:nstep-rmse}: a longer uncorrected return sums the negative advantages of more logged actions, while the gate lets HQL extend a target only through chunks that pass the value comparison.

\subsection{Q3: Combining HQL with Other Policy Extraction Methods}
\label{sec:combine}

\begin{wraptable}{r}{0.5\linewidth}
\centering
\vspace{-1.2em}
\caption{\textbf{HQL as a drop-in critic for different policy extraction methods.} Success rates (\%) on task~2, offline $\to$ online; we bold the better score of each pair. HQL improves both QAM and VGF on all four tasks.}
\label{tab:combine-online}
\vspace{0.2em}
\setlength{\tabcolsep}{2.0pt}
\scriptsize
\begin{tabular}{lcccc}
\toprule
& \texttt{cube-} & \texttt{cube-} & \texttt{antmz-} & \texttt{hmz-} \\
& \texttt{double} & \texttt{triple} & \texttt{giant} & \texttt{giant} \\
\midrule
QAM      & \gv{29}\,$\to$\,\gv{50} & \bv{1}\,$\to$\,\gv{1}
                           & \bv{14}\,$\to$\,\gv{83} & \bv{1}\,$\to$\,\gv{0} \\
\ +\,HQL & \bv{38}\,$\to$\,\bv{99} & \gv{0}\,$\to$\,\bv{80}
                           & \gv{0}\,$\to$\,\bv{98} & \gv{0}\,$\to$\,\bv{63} \\
\midrule
VGF      & \gv{69}\,$\to$\,\gv{74} & \gv{3}\,$\to$\,\gv{25}
                           & \bv{1}\,$\to$\,\gv{28} & \bv{0}\,$\to$\,\gv{0} \\
\ +\,HQL & \bv{80}\,$\to$\,\bv{89} & \bv{4}\,$\to$\,\bv{37}
                           & \gv{0}\,$\to$\,\bv{90} & \bv{0}\,$\to$\,\bv{4} \\
\bottomrule
\end{tabular}
\vspace{-0.8em}
\end{wraptable}
\paragraph{HQL with QAM.} Q-learning with adjoint matching (QAM) \citep{li2026qam} trains a flow policy to sample from the tilted distribution $\pi\propto\mu\exp(\tau Q)$. 
To combine it with HQL, we keep the hard gate unchanged and only replace the best-of-$N$ policy with the QAM policy.
With HQL as its critic, QAM improves on all four tasks at the end of online training, especially on \texttt{cube-triple} ($1\%\to80\%$) and \texttt{humanoidmaze-giant} ($0\%\to63\%$).
\paragraph{HQL with value gradient flow (VGF).} VGF \citep{xu2026vgf} can be thought of an implicit policy that transports from behavior samples toward high-value actions with a few steps of $Q$-gradient ascent. With HQL as its critic, VGF again improves on all four tasks, most on \texttt{antmaze-giant} ($28\%\to90\%$).

\subsection{Q4: Ablations of HQL}
\label{sec:ablations}

\paragraph{Does the window length $H$ matter?} We find the effect is small. At a fixed budget of $512$ targets per update, $H\in\{8,16,32\}$ (batch $64/32/16$) gives similar performance: $0.607$, $0.646$, and $0.621$ (Figure~\ref{fig:ablations}, left). This agrees with the trace length analysis in Section~\ref{sec:benchmark}: since empirically no target used more than six steps, a window of $16$ already exceeds what the gate uses. HQL is therefore not sensitive to $H$ unless it is too small.
\paragraph{Does the best-of-$N$ width matter for HQL?} We try best-of-$N$ width from $\{4,8,16\}$. On OGBench datasets we find that a smaller $N$ is better: $N\in\{4,8,16\}$ gives the mean value of $0.596$, $0.521$, and $0.461$, three times the spread of $H$ (Figure~\ref{fig:ablations}, right). Intuitively, a logged action exchangeable with the $N$ proposals beats all of them with probability $1/(N{+}1)$, so a larger $N$ accepts fewer logged actions and HQL approaches 1-step TD when $N$ is large.
We provide more details on the ablation setup in Appendix~\ref{app:ablations}.

\FloatBarrier
\section{Conclusion and Limitations}
In this paper, we propose HQL, which shortens the effective horizon of Q-learning without inheriting the bias of suboptimal logged actions. HQL gates each logged action with a density-free value comparison and achieves strong results on long-horizon OGBench tasks, both alone and as a drop-in critic for QAM and VGF. We believe that HQL represents a concrete step toward horizon-scalable off-policy RL. One limitation is that the gate relies on the critic's own ranking. One future work is to apply HQL over subgoals in a hierarchical manner.

\section*{Acknowledgement}
We thank Claas Voelcker and members from MIDI lab for discussions on the method and feedback on the early draft of the paper. HX is supported by the Amazon Fellowship. This research used the computational cluster resource provided by the Texas Advanced Computing Center at UT Austin.

\bibliography{refs}

\newpage
\appendix
\FloatBarrier
\section{Appendix}

\FloatBarrier
\subsection{Derivation of the One-Step Error Accumulation}
\label{app:onestep}

This appendix states the residual model behind Eq.~\ref{eq:rmse} and derives the rate. For a family of chains with growing horizon, assume the one-step residuals $e_t=Q_t-r_t-Q_{t+1}$ form a second-order stationary process whose statistics do not depend on $H$:
\begin{equation}
    \mathbb{E}[e_t]=0,
    \qquad
    \operatorname{Var}(e_t)=\sigma^2,
    \qquad
    C(\ell)=\operatorname{Cov}(e_t,e_{t+\ell}),
\end{equation}
with weak correlation in the form $\sum_{\ell=1}^{\infty}|C(\ell)|<\infty$. Taking the mean and variance of the sum in Eq.~\ref{eq:telescope} with $n=H-t$ terms gives
\begin{align}
    \mathbb{E}[\epsilon_t]
    &=0,\\
    \operatorname{Var}(\epsilon_t)
    &=
    n\sigma^2
    +
    2\sum_{\ell=1}^{n-1}(n-\ell)C(\ell).
    \label{eq:variance}
\end{align}
By summability of $|C(\ell)|$, as $n\to\infty$,
\begin{equation}
    \frac{\operatorname{Var}(\epsilon_t)}{n}
    \longrightarrow
    \sigma_{\mathrm{eff}}^2
    :=
    \sigma^2+2\sum_{\ell=1}^{\infty}C(\ell),
\end{equation}
the long-run variance, which is necessarily nonnegative. If $\sigma_{\mathrm{eff}}^2>0$, then $\operatorname{RMSE}(\epsilon_t)\sim\sigma_{\mathrm{eff}}\sqrt{H-t}$, which is Eq.~\ref{eq:rmse}; this is the nondegenerate random-walk regime.

\paragraph{When the rate does not apply.} The random-walk rate is conditional on the residual model and is not a worst case. Negative correlation can give $\sigma_{\mathrm{eff}}^2=0$ and $o(\sqrt{H-t})$ error, while a systematic same-sign residual of horizon-independent size gives $\Theta(H)$. Such a residual arises, for example, when the critic's values are shrunk toward zero under all-negative rewards: the semi-gradient fixed point with shrinkage $\kappa$ satisfies $e_t=-\kappa Q_t>0$ at every step. Accumulation also requires persistent residuals: in a fully expressive tabular chain, exact convergence gives $Q=Q^\pi$, $e_t=0$, and no accumulation, whereas function approximation, finite optimization, and target-network lag all leave residuals.

\paragraph{Discounting.} With $\gamma<1$ the telescoped sum is geometrically weighted, $\epsilon_t=\sum_{k=0}^{H-t-1}\gamma^k e_{t+k}$, and for uncorrelated residuals $\operatorname{Var}(\epsilon_t)=\sigma^2(1-\gamma^{2(H-t)})/(1-\gamma^2)$. The factor grows linearly in $H-t$ until $H-t$ is of order $1/(1-\gamma)$ and saturates at $1/(1-\gamma^2)$ afterwards, so the discount caps the accumulation at its own effective horizon ($\approx100$ steps at $\gamma=0.995$) rather than removing it; the analysis applies with $H$ replaced by $\min\{H,1/(1-\gamma)\}$ up to constants.

\FloatBarrier
\subsection{Derivation of the $n$-Step Error Accumulation}
\label{app:nstep}

\paragraph{Single-target decomposition.} With $A^\pi(s,a)=Q^\pi(s,a)-V^\pi(s)$ and endpoint error $\epsilon_V(s)=\overline V^\pi(s)-V^\pi(s)$, expanding the target of Eq.~\ref{eq:nstep-target} over the behavior continuation gives
\begin{equation}
    \mathbb{E}_\mu
    \left[
        Y_t^{(n)}-Q^\pi(s_t,a_t)
        \,\middle|\,
        s_t,a_t
    \right]
    =
    \sum_{j=1}^{n-1}
    \mathbb{E}_\mu
    \left[
        A^\pi(s_{t+j},a_{t+j})
        \,\middle|\,
        s_t,a_t
    \right]
    +
    \mathbb{E}_\mu
    \left[
        \epsilon_V(s_{t+n})
        \,\middle|\,
        s_t,a_t
    \right].
    \label{eq:nstep-error-decomposition}
\end{equation}
The derivation uses only $r_{t+j}+V^\pi(s_{t+j+1})=Q^\pi(s_{t+j},a_{t+j}) =V^\pi(s_{t+j})+A^\pi(s_{t+j},a_{t+j})$ at each spanned transition (deterministic transitions given the action; the $j=0$ term contributes no advantage because the target evaluates $Q^\pi(s_t,a_t)$ at the logged action itself). The first term is the behavior-action bias inside the segment, the second the endpoint bootstrap error.

\paragraph{Chaining blocks along the horizon.} To turn the per-target decomposition into a start-state statement, partition the chain into $H/n$ blocks of length $n$ (assume $n\mid H$ for clarity) and define the $n$-step training residual $\tilde e_t=Q_t-Y_t^{(n)}$ at each block start. Adding and subtracting $Q^\pi$ and applying Eq.~\ref{eq:nstep-error-decomposition},
\begin{equation}
    \epsilon_t
    =
    \tilde e_t
    +
    b_t(n)
    +
    \epsilon_{t+n},
    \qquad
    b_t(n)=\sum_{j=1}^{n-1}A^\pi(s_{t+j},a_{t+j}),
\end{equation}
where we identify the endpoint state-value error with the action-value error at the next block start (exact on the deterministic evaluation chain). Iterating over blocks with $\epsilon_H=0$,
\begin{equation}
    \epsilon_0
    =
    \sum_{j=0}^{H/n-1}
    \left[
        \tilde e_{jn}+b_{jn}(n)
    \right].
\end{equation}
Model the training residuals $\tilde e_{jn}$ as zero-mean, weakly correlated across blocks with per-block variance $\sigma_n^2=L^{(n)}_{\mathrm{TD}}$ (the analogue of the model in Appendix~\ref{app:onestep}); the bias terms $b_{jn}(n)$, by contrast, share a sign for a near-optimal target policy and add coherently. Then
\begin{equation}
    \operatorname{RMSE}(\epsilon_0)^2
    \approx
    \frac{H}{n}\,\sigma_n^2
    +
    \Bigg(\sum_{j=0}^{H/n-1}b_{jn}(n)\Bigg)^{\!2},
\end{equation}
and since the bias sum covers all logged actions except the $H/n$ block starts, $\sum_j b_{jn}(n)=(1-\tfrac1n)H\bar A_\mu$ with $\bar A_\mu$ the average spanned advantage, which is Eq.~\ref{eq:nstep-rmse}. The result is schematic in the same sense as Appendix~\ref{app:onestep}: it conditions on the residual model, ignores cross terms between residuals and biases, and uses block-aligned targets, but none of these affect the two scalings it exhibits---a $\sqrt{H/n}$ random-walk term and a bias term that grows with $n$ toward the horizon-linear ceiling $H|\bar A_\mu|$.

\FloatBarrier
\subsection{Combination-Lock Dataset}
\label{app:combo-lock-q}
\begin{figure}[!htbp]
    \centering
\includegraphics[width=0.9\linewidth]{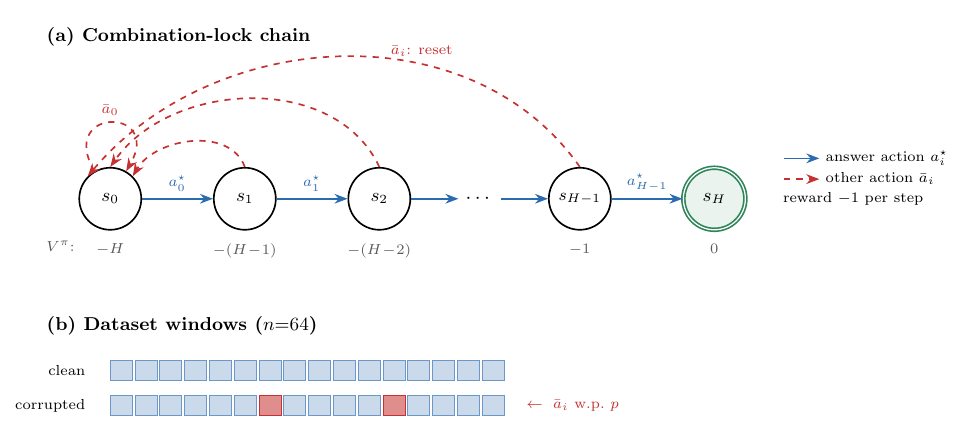}
    \caption{The combination lock dataset with linearly arranged H states.}
    \label{fig:combination-lock}
\end{figure}
In this didactic setup (Figure \ref{fig:combination-lock}), the policy $\pi$ is the deterministic answer policy, and the dynamics are deterministic and acyclic under $\pi$, so the action-value function $Q^{\pi}$ has a closed form. We anchor at the terminal state, $V^\pi(s_H) = 0$. The Bellman equation then reduces to the recursion $V^\pi(s_i) = -1 + V^\pi(s_{i+1})$, which unrolls to
\begin{equation}
    V^\pi(s_i) = -(H-i), \qquad i = 0, \dots, H-1.
\end{equation}
Taking the answer action costs one step and advances the chain. Any other action costs one step and resets to the start. Hence
\begin{align}
    Q^\pi(s_i, a^\star_i) &= -1 + V^\pi(s_{i+1}) = -(H-i), \\
    Q^\pi(s_i, \bar a_i)  &= -1 + V^\pi(s_0) = -(H+1).
\end{align}

\FloatBarrier
\subsection{OGBench Benchmark: Experiment Details and Per-Task Results}
\label{app:h16-setup}

\begin{figure}[!htbp]
\centering
\includegraphics[width=\linewidth]{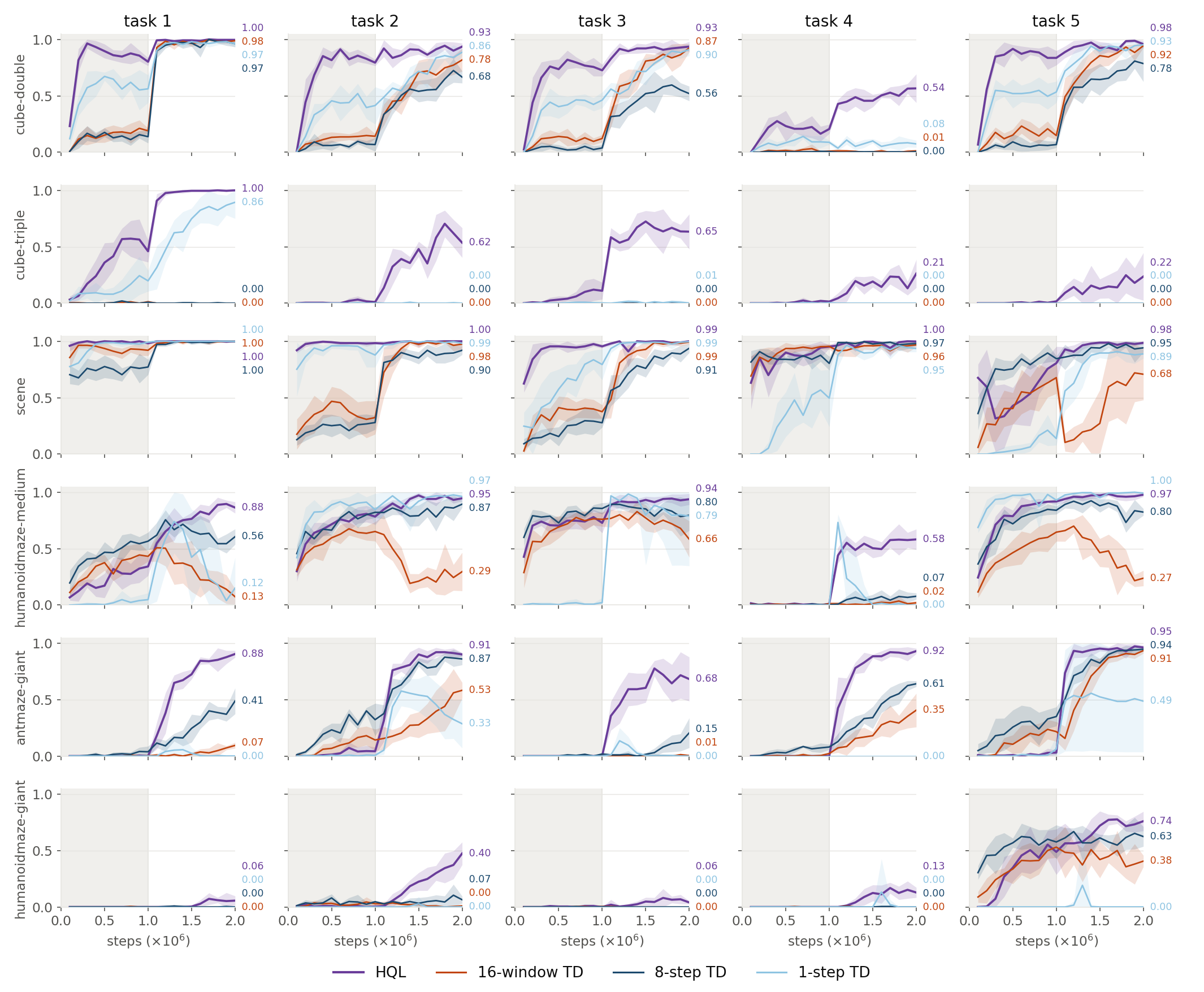}
\caption{\textbf{Per-task view of the OGBench benchmark (Figure~\ref{fig:h16-taskmean}).} Reading down a column shows how one task index behaves across environments, and a row shows how one environment varies over tasks.}
\label{fig:h16-pertask}
\end{figure}

This appendix documents the configuration behind the OGBench benchmark figure of Section~\ref{sec:benchmark} (Figure~\ref{fig:h16-taskmean}) and reports the per-task curves that its task means aggregate over (Figure~\ref{fig:h16-pertask}).

\paragraph{Tasks and datasets.} We use six environment--discount configurations: \texttt{cube-double}, \texttt{cube-triple}, \texttt{scene}, and \texttt{humanoidmaze-medium} at $\gamma=0.99$, and \texttt{antmaze-giant} and \texttt{humanoidmaze-giant} at $\gamma=0.995$; Table~\ref{tab:ogbench-tasks} lists dataset sizes, episode lengths, and dimensionalities. Every group provides five single-task variants that share one dataset and differ only in the reward relabelling, and we run all five, so each panel of Figure~\ref{fig:h16-taskmean} aggregates five tasks. \texttt{scene} uses the sparse reward relabelling.

\paragraph{Training runs.} Each run trains for $10^6$ offline gradient steps followed by $10^6$ online steps, evaluating every $10^5$ steps with 50 episodes; the reported score is the mean success rate over the last three evaluations. Each method is run across the five different tasks corresponding to each of the six OGBench environments for five seeds. Curves show the mean over seeds of the task-averaged score, with bands of
$\pm1$ standard deviation across seeds; we average over tasks before computing seed statistics because per-task success is bimodal, so the across-task spread would
swamp seed variance. Figure~\ref{fig:h16-pertask} gives the per-task
breakdown.

\paragraph{HQL configuration.} HQL uses a window of $H=16$ with 32 windows per batch, i.e., $32 \times 16 = 512$ critic targets per update under the accounting of Section~\ref{sec:hwindow}. The gate is the best-of-$N$ indicator of Eq.~\ref{eq:bon-gate} with $N=4$, computed from the target critic. The policy is a behavior-cloned flow sampler with best-of-$N$ action selection ($N=4$) used both for acting and for the bootstrap $\overline V^\pi$, so the gate, the bootstrap, and the actor all refer to the same target policy. Table~\ref{tab:h16-hparams} lists the full configuration.

\paragraph{Baseline configurations.} The three TD baselines share all networks and
optimization settings with HQL (Table~\ref{tab:h16-hparams}) and differ only in how
their $512$ critic targets per update are formed (Table~\ref{tab:baselines}).

\begin{table}[!htbp]
\centering
\small
\caption{How each method forms its $512$ critic targets per update.}
\label{tab:baselines}
\begin{tabular}{@{}lcccl@{}}
\toprule
Method & Seq.\ length & Seq.\ per update & Targets per seq. & Target \\
\midrule
HQL           & 16 & 32  & 16 & gated, every position \\
16-window TD  & 16 & 32  & 16 & $n$-step, every position \\
8-step TD     & 8  & 512 & 1  & 8-step, first state \\
1-step TD     & 1  & 512 & 1  & 1-step \\
\bottomrule
\end{tabular}
\end{table}
\begin{table}[!htbp]
\centering
\caption{Hyperparameters for the OGBench benchmark experiment (Figures~\ref{fig:h16-taskmean} and~\ref{fig:h16-pertask}).}
\label{tab:h16-hparams}
\begin{tabular}{ll}
\toprule
Parameter & Value \\
\midrule
Window length $H$                       & 16 \\
Windows per batch                       & 32 \\
Critic targets per update               & 512 \quad ($=32\times16$) \\
Gate width $N$ (Eq.~\ref{eq:bon-gate})  & 4 \\
Gate critic                             & target network $\overline Q$ \\
Discount factor                         & 0.99 (0.995 for the two giant mazes) \\
Optimizer                               & Adam \citep{kingma2017adam} \\
Learning rate                           & $3\times10^{-4}$ \\
Target network update rate              & $5\times10^{-3}$ \\
Critic ensemble size                    & 2 \\
Target $Q$ aggregation                  & mean \\
Network width                           & 512 \\
Network depth                           & 4 \\
Activation function                     & GELU \citep{hendrycks2023gelu} \\
Critic layer norm                       & Yes \\
Actor layer norm                        & No \\
Number of flow steps (behavior sampler) & 10 \\
Best-of-$N$ action samples (actor and bootstrap) & 4 \\
Action chunk size                       & 1 (no action chunking) \\
UTD ratio                               & 1 \\
Offline training steps                  & $10^{6}$ \\
Online training steps                   & $10^{6}$ \\
Evaluation interval / episodes          & $10^{5}$ / 50 \\
Seeds per cell                          & 5 (seeds 0--4) \\
\bottomrule
\end{tabular}
\end{table}

\begin{table}[!htbp]
\centering
\caption{OGBench state-based task characteristics. Episode length is the maximum number of environment steps per episode. Rows marked $\dagger$ are the groups used in this work.}
\label{tab:ogbench-tasks}
\begin{tabular}{lrrrrc}
\toprule
Task group & \makecell[r]{Dataset size\\(transitions)} & \makecell[r]{Episode\\length} &
\makecell[r]{Action\\dimension} & \makecell[r]{Observation\\dimension} &
\makecell[c]{Default\\task} \\
\midrule
\multicolumn{6}{l}{\textit{Locomotion} (\texttt{locomaze})} \\
\texttt{pointmaze-medium}      & 1M & 1000 & 2  & 2  & 1 \\
\texttt{pointmaze-large}       & 1M & 1000 & 2  & 2  & 1 \\
\texttt{pointmaze-giant}       & 1M & 1000 & 2  & 2  & 1 \\
\texttt{pointmaze-teleport}    & 1M & 1000 & 2  & 2  & 1 \\
\texttt{antmaze-medium}        & 1M & 1000 & 8  & 29 & 1 \\
\texttt{antmaze-large}         & 1M & 1000 & 8  & 29 & 1 \\
\texttt{antmaze-giant}$^\dagger$ & 1M & 1000 & 8  & 29 & 1 \\
\texttt{antmaze-teleport}      & 1M & 1000 & 8  & 29 & 1 \\
\texttt{humanoidmaze-medium}$^\dagger$ & 2M & 2000 & 21 & 69 & 1 \\
\texttt{humanoidmaze-large}    & 2M & 2000 & 21 & 69 & 1 \\
\texttt{humanoidmaze-giant}$^\dagger$    & 4M & 4000 & 21 & 69 & 1 \\
\texttt{antsoccer-arena}       & 1M & 1000 & 8  & 42 & 4 \\
\texttt{antsoccer-medium}      & 4M & 1000 & 8  & 42 & 4 \\
\midrule
\multicolumn{6}{l}{\textit{Manipulation} (\texttt{manipspace})} \\
\texttt{cube-single}           & 1M & 200  & 5 & 28 & 2 \\
\texttt{cube-double}$^\dagger$ & 1M & 500  & 5 & 37 & 2 \\
\texttt{cube-triple}$^\dagger$ & 3M & 1000 & 5 & 46 & 2 \\
\texttt{cube-quadruple}        & 5M & 1000 & 5 & 55 & 2 \\
\texttt{cube-octuple}          & -- & 1500 & 5 & -- & 2 \\
\texttt{scene}$^\dagger$       & 1M & 750  & 5 & 40 & 2 \\
\texttt{puzzle-3x3}  & 1M & 500  & 5 & 55 & 4 \\
\texttt{puzzle-4x4}            & 1M & 500  & 5 & 83 & 4 \\
\texttt{puzzle-4x5}            & 3M & 1000 & 5 & 99 & 2 \\
\texttt{puzzle-4x6}            & -- & 1000 & 5 & -- & 2 \\
\bottomrule
\end{tabular}
\end{table}

\subsection{Shared Implementation Settings}
\label{app:shared}

Unless a later subsection states otherwise, every experiment uses the networks,
optimizer and behavior flow sampler of Table~\ref{tab:h16-hparams}. All agents are
implemented in one JAX codebase with shared replay, evaluation and logging, so two
compared methods differ only in the settings named for that comparison.

\paragraph{Matched budget.} All comparisons in Section~\ref{sec:benchmark} and the window-length ablation of Section~\ref{sec:ablations} fix the number of critic targets per update at $512$,
following the accounting of \citet{park2025horizon}: a batch of $B$ windows of
length $H$ yields $B\cdot H$ targets for methods that regress every position, and
$B$ targets for methods that regress only the first state. The number of
transitions read per update therefore differs across methods; for example, 8-step
TD reads $4096$ transitions to form its $512$ targets.

\paragraph{Compute.} Runs of $2\times10^6$ steps were executed on 8-GPU A100 and
H100 nodes, eight runs per node, at about $150$ iterations/s per run
($\approx$4--6\,h per run).

\subsection{OCBench Benchmark: Experiment Details}
\label{app:ocbench-setup}

This appendix documents the configuration behind the OCBench results of Section~\ref{sec:benchmark} (Figure~\ref{fig:ocbench}). Unlike the OGBench experiments, we follow the RL baselines of OCBench \citep{park2026ocbench} for every setting other than the critic target, so all three methods share the same networks, critic loss, action space and action chunks (Table~\ref{tab:ocbench-hparams}).

\paragraph{Tasks and datasets.} {\sloppy We use eight full-frequency ($50$\,Hz) tasks, \texttt{block-single}, \texttt{block-double}, \texttt{chamber-easy}, \texttt{chamber-medium}, \texttt{switch-3x3}, \texttt{switch-4x4}, \texttt{hanoi-single} and \texttt{hanoi-double} (all \texttt{task1}), whose episode limits range from $1000$ to $5000$ steps. We train on all episodes of each dataset without filtering for successful ones. Actions are normalized as in OCBench, i.e., the $1$st and $99$th percentiles of each action dimension are mapped to $\pm1$ and the result is clipped at $\pm5$.\par}

\paragraph{Method configurations.} Chunked-HQL uses chunks of $L=25$ actions and windows of $H=8$ chunks ($200$ environment steps) with $128$ windows per batch, i.e., $128\times8=1024$ critic targets per update. Its gate compares the logged chunk with the best of $N=2$ policy chunks, while the bootstrap and acting take the best of $16$ chunks, the same width as the baselines. The $n$-step TD baselines ($n\in\{100,200\}$) regress a single target per sequence with a batch of $1024$. Each run is evaluated with $500$ episodes every $5\times10^4$ steps, and the reported score is the mean success rate over the last three evaluations.

\begin{table}[!htbp]
\centering
\caption{Hyperparameters for the OCBench benchmark experiment (Figure~\ref{fig:ocbench}).}
\label{tab:ocbench-hparams}
\begin{tabular}{ll}
\toprule
Parameter & Value \\
\midrule
Action chunk size $L$                   & 25 (cut at episode boundaries) \\
Window length $H$ (Chunked-HQL)         & 8 chunks ($200$ environment steps) \\
Windows per batch (Chunked-HQL)         & 128 \\
Critic targets per update               & 1024 \quad ($=128\times8$) \\
Batch size of $n$-step TD               & 1024 (one target per sequence) \\
$n$ of $n$-step TD                      & 100, 200 \\
Gate width $N$ (Eq.~\ref{eq:bon-gate})  & 2 \\
Gate critic                             & target network $\overline Q$ \\
Best-of-$N$ (actor and bootstrap)       & 16 \\
Discount factor                         & 0.9999 \\
Optimizer                               & Adam \citep{kingma2017adam} \\
Learning rate                           & $10^{-4}$ \\
Target network update rate              & $10^{-3}$ \\
Critic ensemble size                    & 2 \\
Target $Q$ aggregation                  & mean \\
Network (actor and critic)              & residual MLP, $8\times2048$ \\
Activation function                     & GELU \citep{hendrycks2023gelu} \\
Critic layer norm                       & Yes \\
Actor layer norm                        & Yes \\
Critic loss                             & HL-Gauss, 101 bins on $[0,1]$, $\sigma=0.75$ bins \\
Number of flow steps (behavior sampler) & 10 \\
Flow-time distribution                  & $\mathrm{Beta}(1, 1.5)$ \\
Action normalization                    & percentiles $1/99\mapsto\pm1$, clip at $\pm5$ \\
UTD ratio                               & 1 \\
Offline training steps                  & $10^{6}$ \\
Online training steps                   & $1.5\times10^{6}$ \\
Evaluation interval / episodes          & $5\times10^{4}$ / 500 \\
Seeds per cell                          & 1 (seed 0) \\
\bottomrule
\end{tabular}
\end{table}

\subsection{Soft Gate versus Hard Gate}
\label{app:softgate}

Section~\ref{sec:retrace-connection} shows that the hard gate of Eq.~\ref{eq:bon-gate} is the $\tau\to\infty$ limit of the soft gate of Eq.~\ref{eq:qam-retrace-coefficient}. A natural question is whether a finite $\tau$ works better, since the soft gate keeps partial credit for logged actions that are close to the best candidate. To answer this, we compare the soft gate at $\tau\in\{0,0.5,1,3,30\}$ with HQL's hard gate on six OGBench configurations with five tasks each. Note that $\tau{=}0$ sets every gate to one and recovers uncorrected $8$-step TD. All runs use $H{=}8$, a batch of $128$ windows ($1024$ critic targets per update), $1$M offline and $1$M online steps, and one seed; the soft gate estimates $\log\hat Z$ and acts with $16$ behavior samples, and the hard gate uses $N{=}4$ for both.

\begin{figure}[!htbp]
\centering
\includegraphics[width=\linewidth]{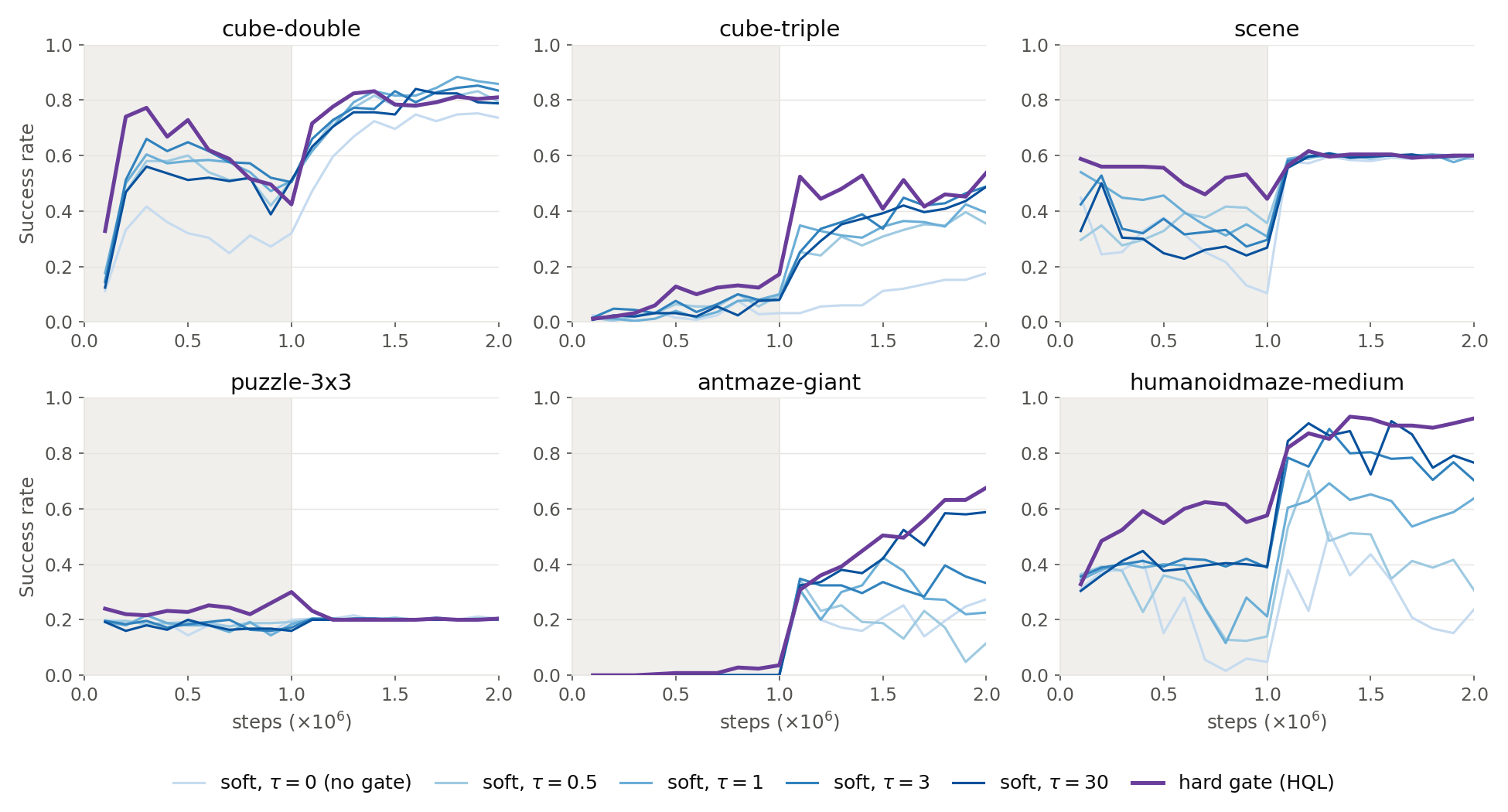}
\caption{\textbf{Soft gate at different temperatures versus HQL's hard gate.} Success rate averaged over the five tasks of each configuration, one seed; gray marks the offline phase. The soft gate improves as $\tau$ grows toward the hard gate, and the hard gate is the best on average, especially on the long-horizon mazes.}
\label{fig:softgate-tau}
\end{figure}

\begin{table}[!htbp]
\centering
\caption{\textbf{Soft gate versus hard gate.} Success rate (\%) at the end of online training, mean of the last three evaluations over five tasks, one seed. We bold the best score in each row at integer level.}
\label{tab:softgate-tau}
\small
\begin{tabular}{lcccccc}
\toprule
& \multicolumn{5}{c}{soft gate} & hard gate \\
\cmidrule(lr){2-6}
& $\tau{=}0$ & $\tau{=}0.5$ & $\tau{=}1$ & $\tau{=}3$ & $\tau{=}30$ & (HQL) \\
\midrule
\texttt{cube-double} & 75 & 81 & \bv{87} & 84 & 80 & 81 \\
\texttt{cube-triple} & 16 & 37 & 39 & 46 & 44 & \bv{48} \\
\texttt{scene} & 59 & \bv{60} & 59 & \bv{60} & \bv{60} & \bv{60} \\
\texttt{puzzle-3x3} & \bv{21} & 20 & 20 & 20 & 20 & 20 \\
\texttt{antmaze-giant} & 24 & 11 & 24 & 36 & 58 & \bv{65} \\
\texttt{hmaze-medium} & 19 & 37 & 60 & 72 & 77 & \bv{91} \\
\midrule
\textbf{mean} & 36 & 41 & 48 & 53 & 57 & \bv{61} \\
\bottomrule
\end{tabular}
\end{table}

Figure~\ref{fig:softgate-tau} and Table~\ref{tab:softgate-tau} show that the hard gate achieves the best mean success rate ($61\%$), and that the soft gate improves steadily as $\tau$ increases ($36\%\to57\%$), i.e., as it approaches the hard gate. The gap is largest on the long-horizon mazes: on \texttt{humanoidmaze-medium} the hard gate reaches $91\%$ against $77\%$ for the best soft gate, and on \texttt{antmaze-giant} $65\%$ against $58\%$. The soft gate is only better on \texttt{cube-double} at $\tau{=}1$, while \texttt{scene} and \texttt{puzzle-3x3} end at the same value for every gate. This indicates that the hard gate is not only the limit that removes $\tau$ as a hyperparameter, but also the strongest member of the family: a finite $\tau$ keeps partial credit for suboptimal logged actions, so the multi-step target inherits part of their bias, which is exactly the failure mode of uncorrected $n$-step TD.

\FloatBarrier
\subsection{Ablation Details}
\label{app:ablations}

Both panels of Figure~\ref{fig:ablations} use HQL with the shared settings of Appendix~\ref{app:shared}. Each panel averages only over the environment--task cells that were run for every value of the ablated setting, so that only that setting varies within a panel.

\paragraph{Window length $H$.} We fix the budget at $512$ critic targets per update, so the number of windows per batch decreases as $H$ increases ($64$, $32$ and $16$ for $H{=}8$, $16$ and $32$ respectively). The panel covers six environment--discount configurations with five tasks each, i.e., $30$ cells. 

\paragraph{Gate width $N$.} We run this panel at $H{=}8$ with $128$ windows per batch, i.e., $1024$ critic targets per update, with two seeds per setting (we use $512$ critic targets everywhere else in the paper). Here $N$ sets both the number of candidates in the gate and the number of best-of-$N$ samples used for acting, so the panel varies the two together. 

\end{document}

%% file: math_commands.tex
\usepackage{amsmath,amsfonts,bm}

\def\eqref#1{equation~\ref{#1}}

\def\1{\bm{1}}

\DeclareMathAlphabet{\mathsfit}{\encodingdefault}{\sfdefault}{m}{sl}
\SetMathAlphabet{\mathsfit}{bold}{\encodingdefault}{\sfdefault}{bx}{n}

%% file: main.bbl
\begin{thebibliography}{74}
\providecommand{\natexlab}[1]{#1}
\providecommand{\url}[1]{\texttt{#1}}
\expandafter\ifx\csname urlstyle\endcsname\relax
  \providecommand{\doi}[1]{doi: #1}\else
  \providecommand{\doi}{doi: \begingroup \urlstyle{rm}\Url}\fi

\bibitem[Abraham et~al.(2026)Abraham, Shi, and Finn]{abraham2026lql}
Armaan~A. Abraham, Lucy~Xiaoyang Shi, and Chelsea Finn.
\newblock Long-horizon {Q}-learning: Accurate value learning via $n$-step inequalities.
\newblock \emph{arXiv preprint arXiv:2605.05812}, 2026.

\bibitem[Ahn et~al.(2025)Ahn, Choi, Han, and Moon]{ahn2025option}
Hongjoon Ahn, Heewoong Choi, Jisu Han, and Taesup Moon.
\newblock Option-aware temporally abstracted value for offline goal-conditioned reinforcement learning.
\newblock In \emph{Advances in Neural Information Processing Systems}, 2025.

\bibitem[Bacon et~al.(2017)Bacon, Harb, and Precup]{bacon2017option}
Pierre-Luc Bacon, Jean Harb, and Doina Precup.
\newblock The option-critic architecture.
\newblock In \emph{AAAI Conference on Artificial Intelligence}, 2017.

\bibitem[Chen et~al.(2023)Chen, Lu, Ying, Su, and Zhu]{chen2023sfbc}
Huayu Chen, Cheng Lu, Chengyang Ying, Hang Su, and Jun Zhu.
\newblock Offline reinforcement learning via high-fidelity generative behavior modeling.
\newblock In \emph{International Conference on Learning Representations (ICLR)}, 2023.

\bibitem[Daley(2026)]{daley2026gated}
Brett Daley.
\newblock Gated {Q}-learning: Add off-policy bias to taste.
\newblock In \emph{Reinforcement Learning Journal}, volume~7, 2026.

\bibitem[Daley et~al.(2023)Daley, White, Amato, and Machado]{daley2023trajectory}
Brett Daley, Martha White, Christopher Amato, and Marlos~C. Machado.
\newblock Trajectory-aware eligibility traces for off-policy reinforcement learning.
\newblock In \emph{International Conference on Machine Learning (ICML)}, 2023.

\bibitem[De~Asis et~al.(2018)De~Asis, Hernandez-Garcia, Holland, and Sutton]{deasis2018multistep}
Kristopher De~Asis, J.~Fernando Hernandez-Garcia, G.~Zacharias Holland, and Richard~S. Sutton.
\newblock Multi-step reinforcement learning: A unifying algorithm.
\newblock In \emph{AAAI Conference on Artificial Intelligence}, 2018.

\bibitem[Espeholt et~al.(2018)Espeholt, Soyer, Munos, Simonyan, Mnih, Ward, Doron, Firoiu, Harley, Dunning, Legg, and Kavukcuoglu]{espeholt2018impala}
Lasse Espeholt, Hubert Soyer, R{\'e}mi Munos, Karen Simonyan, Volodymyr Mnih, Tom Ward, Yotam Doron, Vlad Firoiu, Tim Harley, Iain Dunning, Shane Legg, and Koray Kavukcuoglu.
\newblock {IMPALA}: Scalable distributed deep-{RL} with importance weighted actor-learner architectures.
\newblock In \emph{International Conference on Machine Learning (ICML)}, 2018.

\bibitem[Eysenbach et~al.(2019)Eysenbach, Salakhutdinov, and Levine]{eysenbach2019search}
Benjamin Eysenbach, Ruslan Salakhutdinov, and Sergey Levine.
\newblock Search on the replay buffer: Bridging planning and reinforcement learning.
\newblock In \emph{Advances in Neural Information Processing Systems (NeurIPS)}, 2019.

\bibitem[Farahmand et~al.(2010)Farahmand, Szepesv{\'a}ri, and Munos]{farahmand2010error}
Amir-massoud Farahmand, Csaba Szepesv{\'a}ri, and R{\'e}mi Munos.
\newblock Error propagation for approximate policy and value iteration.
\newblock In \emph{Advances in Neural Information Processing Systems (NeurIPS)}, 2010.

\bibitem[Fujimoto and Gu(2021)]{fujimoto2021minimalist}
Scott Fujimoto and Shixiang~Shane Gu.
\newblock A minimalist approach to offline reinforcement learning.
\newblock In \emph{Advances in Neural Information Processing Systems (NeurIPS)}, 2021.

\bibitem[Fujimoto et~al.(2019)Fujimoto, Meger, and Precup]{fujimoto2019bcq}
Scott Fujimoto, David Meger, and Doina Precup.
\newblock Off-policy deep reinforcement learning without exploration.
\newblock In \emph{International Conference on Machine Learning (ICML)}, 2019.

\bibitem[Hansen-Estruch et~al.(2023)Hansen-Estruch, Kostrikov, Janner, Kuba, and Levine]{hansenestruch2023idql}
Philippe Hansen-Estruch, Ilya Kostrikov, Michael Janner, Jakub~Grudzien Kuba, and Sergey Levine.
\newblock {IDQL}: Implicit {Q}-learning as an actor-critic method with diffusion policies.
\newblock \emph{arXiv preprint arXiv:2304.10573}, 2023.

\bibitem[Harutyunyan et~al.(2016)Harutyunyan, Bellemare, Stepleton, and Munos]{harutyunyan2016qlambda}
Anna Harutyunyan, Marc~G. Bellemare, Tom Stepleton, and R{\'e}mi Munos.
\newblock {$Q(\lambda)$} with off-policy corrections.
\newblock In \emph{International Conference on Algorithmic Learning Theory (ALT)}, 2016.

\bibitem[Hendrycks and Gimpel(2016)]{hendrycks2023gelu}
Dan Hendrycks and Kevin Gimpel.
\newblock Gaussian error linear units ({GELUs}).
\newblock \emph{arXiv preprint arXiv:1606.08415}, 2016.

\bibitem[Hernandez-Garcia and Sutton(2019)]{hernandez2019understanding}
J.~Fernando Hernandez-Garcia and Richard~S. Sutton.
\newblock Understanding multi-step deep reinforcement learning: A systematic study of the dqn target.
\newblock \emph{arXiv preprint arXiv:1901.07510}, 2019.

\bibitem[Hessel et~al.(2018)Hessel, Modayil, van Hasselt, Schaul, Ostrovski, Dabney, Horgan, Piot, Azar, and Silver]{hessel2018rainbow}
Matteo Hessel, Joseph Modayil, Hado van Hasselt, Tom Schaul, Georg Ostrovski, Will Dabney, Dan Horgan, Bilal Piot, Mohammad Azar, and David Silver.
\newblock Rainbow: Combining improvements in deep reinforcement learning.
\newblock In \emph{AAAI Conference on Artificial Intelligence}, 2018.

\bibitem[Kalashnikov et~al.(2018)Kalashnikov, Irpan, Pastor, Ibarz, Herzog, Jang, Quillen, Holly, Kalakrishnan, Vanhoucke, and Levine]{kalashnikov2018qtopt}
Dmitry Kalashnikov, Alex Irpan, Peter Pastor, Julian Ibarz, Alexander Herzog, Eric Jang, Deirdre Quillen, Ethan Holly, Mrinal Kalakrishnan, Vincent Vanhoucke, and Sergey Levine.
\newblock {QT-Opt}: Scalable deep reinforcement learning for vision-based robotic manipulation.
\newblock In \emph{Conference on Robot Learning (CoRL)}, 2018.

\bibitem[Kapturowski et~al.(2019)Kapturowski, Ostrovski, Quan, Munos, and Dabney]{kapturowski2019r2d2}
Steven Kapturowski, Georg Ostrovski, John Quan, R{\'e}mi Munos, and Will Dabney.
\newblock Recurrent experience replay in distributed reinforcement learning.
\newblock In \emph{International Conference on Learning Representations (ICLR)}, 2019.

\bibitem[Kearns and Singh(2000)]{kearns2000bias}
Michael Kearns and Satinder Singh.
\newblock Bias-variance error bounds for temporal difference updates.
\newblock In \emph{Conference on Learning Theory (COLT)}, 2000.

\bibitem[Kim and hwan Oh(2026)]{kim2026peng}
Byeongchan Kim and Min hwan Oh.
\newblock {Peng}'s q($\lambda$) for conservative value estimation in offline reinforcement learning.
\newblock In \emph{International Conference on Learning Representations (ICLR)}, 2026.

\bibitem[Kingma and Ba(2014)]{kingma2017adam}
Diederik~P. Kingma and Jimmy Ba.
\newblock Adam: A method for stochastic optimization.
\newblock \emph{arXiv preprint arXiv:1412.6980}, 2014.

\bibitem[Kostrikov et~al.(2022)Kostrikov, Nair, and Levine]{kostrikov2022iql}
Ilya Kostrikov, Ashvin Nair, and Sergey Levine.
\newblock Offline reinforcement learning with implicit {Q}-learning.
\newblock In \emph{International Conference on Learning Representations (ICLR)}, 2022.

\bibitem[Kozuno et~al.(2021)Kozuno, Tang, Rowland, Munos, Kapturowski, Dabney, Valko, and Abel]{kozuno2021peng}
Tadashi Kozuno, Yunhao Tang, Mark Rowland, R{\'e}mi Munos, Steven Kapturowski, Will Dabney, Michal Valko, and David Abel.
\newblock Revisiting {Peng's} {Q}($\lambda$) for modern reinforcement learning.
\newblock In \emph{International Conference on Machine Learning (ICML)}, 2021.

\bibitem[Kumar et~al.(2019)Kumar, Fu, Tucker, and Levine]{kumar2019bear}
Aviral Kumar, Justin Fu, George Tucker, and Sergey Levine.
\newblock Stabilizing off-policy {Q}-learning via bootstrapping error reduction.
\newblock In \emph{Advances in Neural Information Processing Systems (NeurIPS)}, 2019.

\bibitem[Kumar et~al.(2020)Kumar, Zhou, Tucker, and Levine]{kumar2020cql}
Aviral Kumar, Aurick Zhou, George Tucker, and Sergey Levine.
\newblock Conservative {Q}-learning for offline reinforcement learning.
\newblock In \emph{Advances in Neural Information Processing Systems (NeurIPS)}, 2020.

\bibitem[Laidlaw et~al.(2023)Laidlaw, Russell, and Dragan]{laidlaw2023effective}
Cassidy Laidlaw, Stuart Russell, and Anca Dragan.
\newblock Bridging {RL} theory and practice with the effective horizon.
\newblock In \emph{Advances in Neural Information Processing Systems (NeurIPS)}, 2023.

\bibitem[Levine et~al.(2016)Levine, Finn, Darrell, and Abbeel]{levine2016end}
Sergey Levine, Chelsea Finn, Trevor Darrell, and Pieter Abbeel.
\newblock End-to-end training of deep visuomotor policies.
\newblock \emph{Journal of Machine Learning Research}, 17\penalty0 (39):\penalty0 1--40, 2016.

\bibitem[Li and Levine(2026)]{li2026qam}
Qiyang Li and Sergey Levine.
\newblock {Q}-learning with adjoint matching.
\newblock \emph{arXiv preprint arXiv:2601.14234}, 2026.

\bibitem[Li et~al.(2025{\natexlab{a}})Li, Park, and Levine]{li2025decoupled}
Qiyang Li, Seohong Park, and Sergey Levine.
\newblock Decoupled {Q-Chunking}.
\newblock \emph{arXiv preprint arXiv:2512.10926}, 2025{\natexlab{a}}.

\bibitem[Li et~al.(2025{\natexlab{b}})Li, Zhou, and Levine]{li2025qc}
Qiyang Li, Zhiyuan Zhou, and Sergey Levine.
\newblock Reinforcement learning with action chunking.
\newblock In \emph{Advances in Neural Information Processing Systems}, 2025{\natexlab{b}}.

\bibitem[Liu et~al.(2018)Liu, Li, Tang, and Zhou]{liu2018breaking}
Qiang Liu, Lihong Li, Ziyang Tang, and Dengyong Zhou.
\newblock Breaking the curse of horizon: Infinite-horizon off-policy estimation.
\newblock In \emph{Advances in Neural Information Processing Systems (NeurIPS)}, 2018.

\bibitem[Mahmood et~al.(2017)Mahmood, Yu, and Sutton]{mahmood2017abq}
Ashique~Rupam Mahmood, Huizhen Yu, and Richard~S. Sutton.
\newblock Multi-step off-policy learning without importance sampling ratios.
\newblock \emph{arXiv preprint arXiv:1702.03006}, 2017.

\bibitem[Mao et~al.(2024{\natexlab{a}})Mao, Xu, Zhan, Zhang, and Zhang]{mao2024diffusiondice}
Liyuan Mao, Haoran Xu, Xianyuan Zhan, Weinan Zhang, and Amy Zhang.
\newblock {Diffusion-DICE}: In-sample diffusion guidance for offline reinforcement learning.
\newblock In \emph{Advances in Neural Information Processing Systems (NeurIPS)}, 2024{\natexlab{a}}.

\bibitem[Mao et~al.(2024{\natexlab{b}})Mao, Xu, Zhang, and Zhan]{mao2024odice}
Liyuan Mao, Haoran Xu, Weinan Zhang, and Xianyuan Zhan.
\newblock {ODICE}: Revealing the mystery of distribution correction estimation via orthogonal-gradient update.
\newblock In \emph{International Conference on Learning Representations (ICLR)}, 2024{\natexlab{b}}.

\bibitem[Munos(2005)]{munos2005error}
R{\'e}mi Munos.
\newblock Error bounds for approximate value iteration.
\newblock In \emph{AAAI Conference on Artificial Intelligence}, 2005.

\bibitem[Munos et~al.(2016)Munos, Stepleton, Harutyunyan, and Bellemare]{munos2016retrace}
R{\'e}mi Munos, Tom Stepleton, Anna Harutyunyan, and Marc~G. Bellemare.
\newblock Safe and efficient off-policy reinforcement learning.
\newblock In \emph{Advances in Neural Information Processing Systems (NeurIPS)}, 2016.

\bibitem[Nachum et~al.(2018)Nachum, Gu, Lee, and Levine]{nachum2018hiro}
Ofir Nachum, Shixiang Gu, Honglak Lee, and Sergey Levine.
\newblock Data-efficient hierarchical reinforcement learning.
\newblock In \emph{Advances in Neural Information Processing Systems (NeurIPS)}, 2018.

\bibitem[Nachum et~al.(2019)Nachum, Tang, Lu, Gu, Lee, and Levine]{nachum2019why}
Ofir Nachum, Haoran Tang, Xingyu Lu, Shixiang Gu, Honglak Lee, and Sergey Levine.
\newblock Why does hierarchy (sometimes) work so well in reinforcement learning?
\newblock \emph{arXiv preprint arXiv:1909.10618}, 2019.

\bibitem[Nair et~al.(2020)Nair, Gupta, Dalal, and Levine]{nair2020awac}
Ashvin Nair, Abhishek Gupta, Murtaza Dalal, and Sergey Levine.
\newblock {AWAC}: Accelerating online reinforcement learning with offline datasets.
\newblock \emph{arXiv preprint arXiv:2006.09359}, 2020.

\bibitem[Park and Levine(2026)]{park2026ocbench}
Seohong Park and Sergey Levine.
\newblock Behavioral cloning mystery.
\newblock \emph{arXiv preprint arXiv:2610.07056}, 2026.

\bibitem[Park et~al.(2023)Park, Ghosh, Eysenbach, and Levine]{park2023hiql}
Seohong Park, Dibya Ghosh, Benjamin Eysenbach, and Sergey Levine.
\newblock {HIQL}: Offline goal-conditioned {RL} with latent states as actions.
\newblock In \emph{Advances in Neural Information Processing Systems (NeurIPS)}, 2023.

\bibitem[Park et~al.(2025{\natexlab{a}})Park, Frans, Eysenbach, and Levine]{park2025ogbench}
Seohong Park, Kevin Frans, Benjamin Eysenbach, and Sergey Levine.
\newblock {OGBench}: Benchmarking offline goal-conditioned {RL}.
\newblock In \emph{International Conference on Learning Representations (ICLR)}, 2025{\natexlab{a}}.

\bibitem[Park et~al.(2025{\natexlab{b}})Park, Frans, Mann, Eysenbach, Kumar, and Levine]{park2025horizon}
Seohong Park, Kevin Frans, Deepinder Mann, Benjamin Eysenbach, Aviral Kumar, and Sergey Levine.
\newblock Horizon reduction makes {RL} scalable.
\newblock In \emph{Advances in Neural Information Processing Systems}, 2025{\natexlab{b}}.

\bibitem[Park et~al.(2025{\natexlab{c}})Park, Li, and Levine]{park2025fql}
Seohong Park, Qiyang Li, and Sergey Levine.
\newblock Flow {Q}-learning.
\newblock In \emph{International Conference on Machine Learning (ICML)}, 2025{\natexlab{c}}.

\bibitem[Park et~al.(2026)Park, Mann, and Levine]{park2025dual}
Seohong Park, Deepinder Mann, and Sergey Levine.
\newblock Dual goal representations.
\newblock In \emph{International Conference on Learning Representations (ICLR)}, 2026.

\bibitem[Peng and Williams(1996)]{peng1996incremental}
Jing Peng and Ronald~J. Williams.
\newblock Incremental multi-step {Q}-learning.
\newblock \emph{Machine Learning}, 22:\penalty0 283--290, 1996.

\bibitem[Peng et~al.(2019)Peng, Kumar, Zhang, and Levine]{peng2019awr}
Xue~Bin Peng, Aviral Kumar, Grace Zhang, and Sergey Levine.
\newblock Advantage-weighted regression: Simple and scalable off-policy reinforcement learning.
\newblock \emph{arXiv preprint arXiv:1910.00177}, 2019.

\bibitem[Peters et~al.(2010)Peters, M{\"u}lling, and Altun]{peters2010reps}
Jan Peters, Katharina M{\"u}lling, and Yasemin Altun.
\newblock Relative entropy policy search.
\newblock In \emph{AAAI Conference on Artificial Intelligence}, 2010.

\bibitem[Precup et~al.(2000)Precup, Sutton, and Singh]{precup2000eligibility}
Doina Precup, Richard~S. Sutton, and Satinder Singh.
\newblock Eligibility traces for off-policy policy evaluation.
\newblock In \emph{International Conference on Machine Learning (ICML)}, 2000.

\bibitem[Rowland et~al.(2020)Rowland, Dabney, and Munos]{rowland2020adaptive}
Mark Rowland, Will Dabney, and R{\'e}mi Munos.
\newblock Adaptive trade-offs in off-policy learning.
\newblock In \emph{International Conference on Artificial Intelligence and Statistics (AISTATS)}, 2020.

\bibitem[Seo and Abbeel(2024)]{seo2025chunk}
Younggyo Seo and Pieter Abbeel.
\newblock Coarse-to-fine {Q}-network with action sequence for data-efficient reinforcement learning.
\newblock \emph{arXiv preprint arXiv:2411.12155}, 2024.

\bibitem[Song et~al.(2026)Song, Park, and Lee]{song2026chunk}
Gwanwoo Song, Kwanyoung Park, and Youngwoon Lee.
\newblock Chunk-guided {Q-Learning}.
\newblock \emph{arXiv preprint arXiv:2603.13971}, 2026.

\bibitem[Sutton and Barto(2018)]{sutton2018book}
Richard~S. Sutton and Andrew~G. Barto.
\newblock \emph{Reinforcement Learning: An Introduction}.
\newblock MIT Press, 2nd edition, 2018.

\bibitem[Sutton et~al.(2016)Sutton, Mahmood, and White]{sutton2016emphatic}
Richard~S. Sutton, A.~Rupam Mahmood, and Martha White.
\newblock An emphatic approach to the problem of off-policy temporal-difference learning.
\newblock \emph{Journal of Machine Learning Research}, 2016.

\bibitem[Touati et~al.(2018)Touati, Bacon, Precup, and Vincent]{touati2018convergent}
Ahmed Touati, Pierre-Luc Bacon, Doina Precup, and Pascal Vincent.
\newblock Convergent tree backup and {Retrace} with function approximation.
\newblock In \emph{International Conference on Machine Learning (ICML)}, 2018.

\bibitem[Tsitsiklis and Roy(1997)]{tsitsiklis1997analysis}
John~N. Tsitsiklis and Benjamin~Van Roy.
\newblock An analysis of temporal-difference learning with function approximation.
\newblock \emph{IEEE Transactions on Automatic Control}, 42\penalty0 (5):\penalty0 674--690, 1997.

\bibitem[van Hasselt et~al.(2018)van Hasselt, Doron, Strub, Hessel, Sonnerat, and Modayil]{vanhasselt2018deadly}
Hado van Hasselt, Yotam Doron, Florian Strub, Matteo Hessel, Nicolas Sonnerat, and Joseph Modayil.
\newblock Deep reinforcement learning and the deadly triad.
\newblock \emph{arXiv preprint arXiv:1812.02648}, 2018.

\bibitem[Vezhnevets et~al.(2017)Vezhnevets, Osindero, Schaul, Heess, Jaderberg, Silver, and Kavukcuoglu]{vezhnevets2017feudal}
Alexander~Sasha Vezhnevets, Simon Osindero, Tom Schaul, Nicolas Heess, Max Jaderberg, David Silver, and Koray Kavukcuoglu.
\newblock {FeUdal} networks for hierarchical reinforcement learning.
\newblock In \emph{International Conference on Machine Learning (ICML)}, 2017.

\bibitem[Wang et~al.(2023{\natexlab{a}})Wang, Torralba, Isola, and Zhang]{wang2023quasimetric}
Tongzhou Wang, Antonio Torralba, Phillip Isola, and Amy Zhang.
\newblock Optimal goal-reaching reinforcement learning via quasimetric learning.
\newblock In \emph{International Conference on Machine Learning (ICML)}, 2023{\natexlab{a}}.

\bibitem[Wang et~al.(2023{\natexlab{b}})Wang, Hunt, and Zhou]{wang2023diffusionql}
Zhendong Wang, Jonathan~J. Hunt, and Mingyuan Zhou.
\newblock Diffusion policies as an expressive policy class for offline reinforcement learning.
\newblock In \emph{International Conference on Learning Representations (ICLR)}, 2023{\natexlab{b}}.

\bibitem[Watkins(1989)]{watkins1989learning}
Christopher J. C.~H. Watkins.
\newblock \emph{Learning from Delayed Rewards}.
\newblock PhD thesis, University of Cambridge, 1989.

\bibitem[White and White(2016)]{white2016greedy}
Martha White and Adam White.
\newblock A greedy approach to adapting the trace parameter for temporal difference learning.
\newblock In \emph{International Conference on Autonomous Agents and Multiagent Systems (AAMAS)}, 2016.

\bibitem[Wu et~al.(2019)Wu, Tucker, and Nachum]{wu2019brac}
Yifan Wu, George Tucker, and Ofir Nachum.
\newblock Behavior regularized offline reinforcement learning.
\newblock \emph{arXiv preprint arXiv:1911.11361}, 2019.

\bibitem[Xu et~al.(2021)Xu, Zhan, Li, and Yin]{xu2021sbac}
Haoran Xu, Xianyuan Zhan, Jianxiong Li, and Honglei Yin.
\newblock Offline reinforcement learning with soft behavior regularization.
\newblock In \emph{NeurIPS Offline Reinforcement Learning Workshop}, 2021.

\bibitem[Xu et~al.(2022)Xu, Jiang, Li, and Zhan]{xu2022por}
Haoran Xu, Li~Jiang, Jianxiong Li, and Xianyuan Zhan.
\newblock A policy-guided imitation approach for offline reinforcement learning.
\newblock In \emph{Advances in Neural Information Processing Systems (NeurIPS)}, 2022.

\bibitem[Xu et~al.(2023)Xu, Jiang, Li, Yang, Wang, Chan, and Zhan]{xu2023ivr}
Haoran Xu, Li~Jiang, Jianxiong Li, Zhuoran Yang, Zhaoran Wang, Victor Wai~Kin Chan, and Xianyuan Zhan.
\newblock Offline {RL} with no {OOD} actions: In-sample learning via implicit value regularization.
\newblock In \emph{International Conference on Learning Representations (ICLR)}, 2023.

\bibitem[Xu et~al.(2025{\natexlab{a}})Xu, Li, Sikchi, Niekum, and Zhang]{xu2025idrl}
Haoran Xu, Shuozhe Li, Harshit Sikchi, Scott Niekum, and Amy Zhang.
\newblock An optimal discriminator weighted imitation perspective for reinforcement learning.
\newblock In \emph{International Conference on Learning Representations (ICLR)}, 2025{\natexlab{a}}.

\bibitem[Xu et~al.(2025{\natexlab{b}})Xu, Mao, Jin, Zhang, Zhan, and Zhang]{xu2025unirl}
Haoran Xu, Liyuan Mao, Hui Jin, Weinan Zhang, Xianyuan Zhan, and Amy Zhang.
\newblock {Uni-RL}: Unifying online and offline {RL} via implicit value regularization.
\newblock In \emph{Advances in Neural Information Processing Systems (NeurIPS)}, 2025{\natexlab{b}}.

\bibitem[Xu et~al.(2026)Xu, Hu, Sojoudi, and Zhang]{xu2026vgf}
Haoran Xu, Kaiwen Hu, Somayeh Sojoudi, and Amy Zhang.
\newblock Reinforcement learning via value gradient flow.
\newblock In \emph{International Conference on Learning Representations (ICLR)}, 2026.

\bibitem[Zhan et~al.(2022)Zhan, Xu, Zhang, Zhu, Yin, and Zheng]{zhan2022deepthermal}
Xianyuan Zhan, Haoran Xu, Yue Zhang, Xiangyu Zhu, Honglei Yin, and Yu~Zheng.
\newblock {DeepThermal}: Combustion optimization for thermal power generating units using offline reinforcement learning.
\newblock In \emph{AAAI Conference on Artificial Intelligence}, 2022.

\bibitem[Zhao et~al.(2023)Zhao, Kumar, Levine, and Finn]{zhao2023act}
Tony~Z. Zhao, Vikash Kumar, Sergey Levine, and Chelsea Finn.
\newblock Learning fine-grained bimanual manipulation with low-cost hardware.
\newblock In \emph{Robotics: Science and Systems (RSS)}, 2023.

\bibitem[Zheng et~al.(2026)Zheng, Myers, Eysenbach, and Levine]{zheng2025multistep}
Bill~Chunyuan Zheng, Vivek Myers, Benjamin Eysenbach, and Sergey Levine.
\newblock Multistep quasimetric learning for scalable goal-conditioned reinforcement learning.
\newblock In \emph{International Conference on Learning Representations}, 2026.

\bibitem[Zhou and Kao(2025)]{zhou2025flattening}
John~L. Zhou and Jonathan~C. Kao.
\newblock Flattening hierarchies with policy bootstrapping.
\newblock In \emph{Advances in Neural Information Processing Systems (NeurIPS)}, 2025.

\end{thebibliography}
